\documentclass{article}
\usepackage{microtype}
\usepackage{graphicx}
\usepackage{subfigure}
\usepackage{booktabs} 
\usepackage{multirow}
\usepackage{makecell}
\usepackage{enumitem}
\usepackage[unicode]{hyperref}

\usepackage[preprint]{icml2026}
\usepackage{amsmath}
\usepackage{amssymb}
\usepackage{mathtools}
\usepackage{amsthm}
\usepackage[capitalize,noabbrev]{cleveref}
\usepackage{CJKutf8} 
\theoremstyle{plain}

\theoremstyle{definition}

\theoremstyle{remark}

\usepackage[textsize=tiny]{todonotes}

\newcommand{\TO}{\textbf{to}\ }
\newcommand{\RETURN}{\STATE \textbf{return}\ }

\icmltitlerunning{From Tokens to Blocks: A Block-Diffusion Perspective on Molecular Generation}

\begin{document}
\begin{CJK*}{UTF8}{gbsn}

\icmlsetsymbol{equal}{*}

\twocolumn[
\icmltitle{From Tokens to Blocks: A Block-Diffusion Perspective on Molecular Generation}

\begin{icmlauthorlist}
\icmlauthor{Qianwei Yang}{szu,equal}
\icmlauthor{Dong Xu}{szu,equal}
\icmlauthor{Zhangfan Yang}{unnc}
\icmlauthor{Sisi Yuan}{szu}
\icmlauthor{Zexuan Zhu}{szu}
\icmlauthor{Jianqiang Li}{szu}
\icmlauthor{Junkai Ji}{szu}
\end{icmlauthorlist}

\icmlaffiliation{szu}{School of Artificial Intelligence, Shenzhen University, Shenzhen 518060, China}
\icmlaffiliation{unnc}{School of Computer Science, University of Nottingham Ningbo, Ningbo 315100, China}

\icmlcorrespondingauthor{Junkai Ji}{jijunkai@szu.edu.cn}

\vskip 0.3in
]

\printAffiliationsAndNotice{\icmlEqualContribution} 

\begin{abstract}
Drug discovery can be viewed as a combinatorial search over an immense chemical space, motivating the development of deep generative models for \textit{de novo} molecular design. Among these, GPT-based molecular language models (MLM) have shown strong molecular design performance by learning chemical syntax and semantics from large-scale data. However, existing MLMs face two fundamental limitations: they inadequately capture the graph-structured nature of molecules when formulated as next-token prediction problems, and they typically lack explicit mechanisms for target-aware generation. Here, we propose SoftMol, a unified framework that co-designs molecular representation, model architecture, and search strategy for target-aware molecular generation. SoftMol introduces soft fragments, a rule-free block representation of SMILES that enables diffusion-native modeling, and develops SoftBD, the first block-diffusion molecular language model that combines local bidirectional diffusion with autoregressive generation under molecular structural constraints. To favor generated molecules with high drug-likeness and synthetic accessibility, SoftBD is trained on a carefully curated dataset named ZINC-Curated. SoftMol further integrates a gated Monte Carlo tree search to assemble fragments in a target-aware manner. Experimental results show that, compared with current state-of-the-art models, SoftMol achieves 100\% chemical validity, improves binding affinity by 9.7\%, yields a 2–3× increase in molecular diversity, and delivers a 6.6× speedup in inference efficiency. Our code is available at \url{https://github.com/SZU-ADDG/SoftMol}.

\end{abstract}

\begin{figure*}[t]
    \centering
    \includegraphics[width=\textwidth]{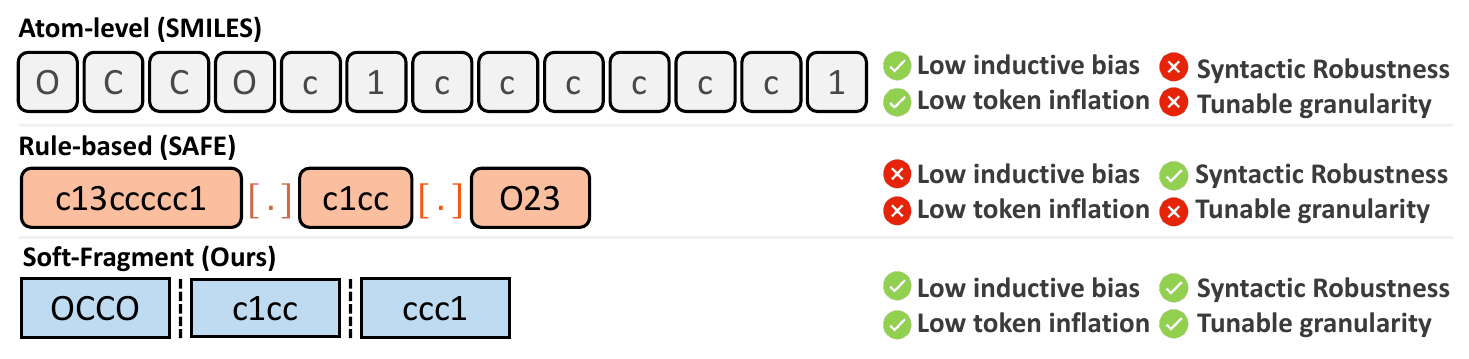}
    \vspace{-4mm}
    \caption{\textbf{Comparison of molecular representation paradigms.} \textbf{Top:} Atom-level SMILES provides minimal bias but lacks syntactic robustness. \textbf{Middle:} Rule-based fragments impose rigid vocabulary constraints and heuristic priors. \textbf{Bottom:} Our soft-fragment approach segments SMILES into fixed-length blocks, enabling tunable granularity and high robustness without heuristic rules or auxiliary tokens.}
    \label{fig:paradigms}
    \vspace{-2mm}
\end{figure*}

\section{Introduction}
\label{sec:intro}
Drug discovery can be formulated as a search problem over a chemical space containing up to $10^{60}$ possible molecules~\citep{polishchuk2013, reymond2015chemical}.
To address the scale of this problem, prior work has increasingly shifted from high-throughput screening toward \textit{de novo} molecular design based on deep generative models~\citep{chen2018rise, vamathevan2019applications, bilodeau2022generative}. A variety of generative architectures have been developed for molecular generation, encompassing autoregressive Transformers~\citep{molgpt, safe2024} alongside diffusion and flow-based models~\citep{ho2020denoising, austin2021structured, genmol}.

GPT-based molecular language models (MLM) have attracted considerable attention among these models, due to their strong empirical performance~\citep{molgpt, flam2022language}. They represent small molecules as textual sequences, such as SMILES and SAFE strings \citep{weininger1988smiles, safe2024}. By training on large-scale molecular datasets, these models learn the intrinsic syntactic and semantic properties of chemical structures, which enables the generation of chemically plausible molecules with high diversity and novelty \citep{selfies2020, nigam2022group_selfies}. 

However, these MLM-based approaches suffer from two fundamental limitations. First, small molecules are intrinsically not one-dimensional and unidirectional sequences but structured chemical graphs~\citep{jin2018jtvae, de2018molgan}. Consequently, framing molecular generation as a next-token prediction problem is suboptimal, as it fails to adequately capture the local chemical context, including the interactions among neighboring atoms within molecular substructures~\citep{atz2021geometric, vignac2023}. Second, such generative models typically operate independently of the target protein, as the generation process cannot explicitly incorporate target-specific information~\citep{bilodeau2022generative, peng2022pocket}. As a result, these limitations render MLMs ill-suited for true target-aware drug design and substantially constrain their practical utility in realistic drug discovery workflows. 

To address the aforementioned limitations, several molecular language models have explored discrete diffusion architectures to explicitly model interactions among all tokens~\citep{genmol, austin2021structured}. However, diffusion-based models are designed to generate output sequences of arbitrary length~\citep{ssdlm, mdlm}. In contrast, small molecules typically contain a bounded and highly structured number of tokens, ranging from only a few to several hundred~\citep{sterling2015zinc}. This mismatch conflicts with the intrinsic structural constraints of molecular representations~\citep{kusner2017gvae, vignac2023}. On the other hand, protein information has been used as prompt or conditional input to guide molecular language models toward target-specific molecule generation~\citep{grechishnikova2021transformer}. However, cross-modal alignment is challenging, and experimental protein–ligand data remain scarce~\citep{huang2021therapeutics}. Together, these issues still limit the scalability and practical applicability of current MLMs for drug design.

This study proposes \textbf{SoftMol}, a unified framework that co-designs molecular representations, model architecture, and search strategy to systematically address these limitations. Specifically, SoftMol partitions SMILES sequences into contiguous, fixed-length blocks without chemistry-specific heuristics or auxiliary tokens, which we term \textbf{soft-fragments}, distinguishing them from traditional rule-based fragments. A block-diffusion–based masked language model, termed \textbf{SoftBD}, is then employed for molecular generation, enabling local bidirectional diffusion within each soft fragment while autoregressively conditioning on previously generated fragments. SoftBD is trained on a carefully curated, high-quality molecular dataset, termed ZINC-Curated, to promote the generation of molecules with strong drug-likeness and synthetic accessibility. Finally, SoftMol integrates a novel gated Monte Carlo tree search (MCTS) procedure to assemble the generated soft fragments in a target-aware manner, facilitating molecular design toward specific protein targets. Extensive experiments demonstrate that SoftMol significantly outperforms all existing molecular generative models on both \textit{de novo} molecule generation and target-specific design tasks, establishing a new state of the art and revealing the decisive advantages of block-diffusion–based modeling for molecular generation.

Our main contributions can be summarized as follows:
\begin{itemize}[leftmargin=*, nosep]
    \item We introduce \textbf{soft-fragments}, a rule-free block representation of SMILES that enables diffusion-native molecular modeling and tunable granularity across tasks.   
    \item We propose \textbf{SoftBD}, the first block-diffusion molecular language model that reconciles bidirectional diffusion with autoregressive generation under molecular structural constraints. 
    \item We develop \textbf{SoftMol}, a unified and target-aware framework that integrates SoftBD and gated MCTS, achieving state-of-the-art performance in both \textit{de novo} and target-specific molecular design.
\end{itemize}

\begin{figure*}[t]
    \centering
    \includegraphics[width=\textwidth]{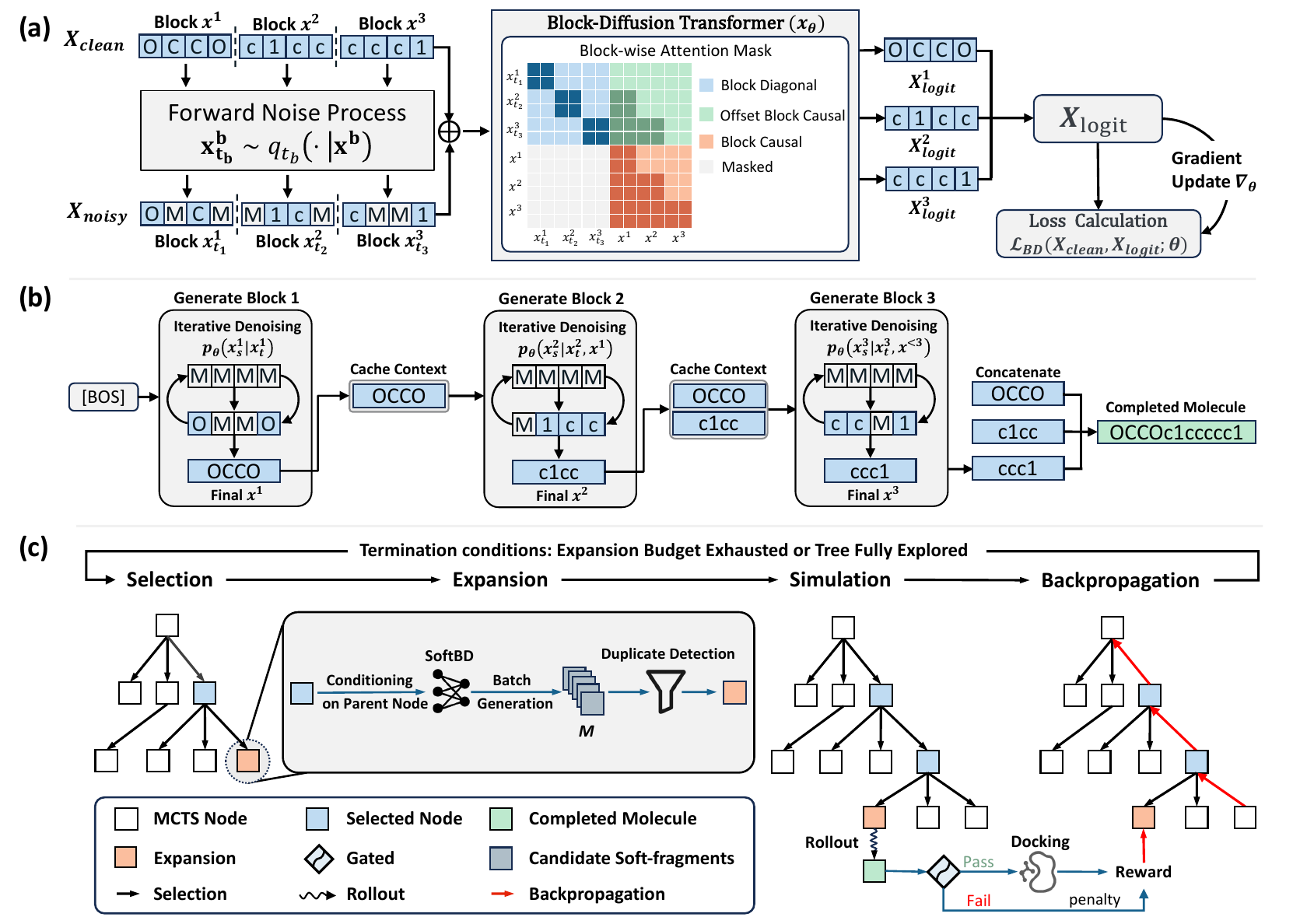}
    \vspace{-4mm}
    \caption{\textbf{Overview of the SoftMol framework.}
    (a) \textbf{Training:} The concatenated clean and noised sequences are processed by the Block-Diffusion Transformer with a block-wise
attention mask enforcing intra-block bidirectional and inter-block causal dependencies.
    (b) \textbf{Sampling:} Starting from \texttt{[BOS]}, molecules are generated semi-autoregressively via iterative denoising, with
previously decoded blocks cached as context.
    (c) \textbf{Gated MCTS:} Selection traverses the tree via UCT; Expansion uses batched SoftBD generation with duplicate filtering;
Simulation applies a tunable feasibility gate---candidates satisfying pharmacological constraints proceed to docking, while failures receive a
penalty; Backpropagation updates node statistics.}
    \vspace{-4mm}
    \label{fig:overview}
\end{figure*}

\section{Preliminaries}
\label{sec:preliminaries}

\subsection{Notation and Problem Setup}
\label{sec:notation}

\textbf{Molecular sequence representation.}
A molecule is represented as a discrete sequence $\mathbf{x} = (x_1, \ldots, x_L)$ of length $L$, where each token $x_i \in \mathcal{V}$ belongs to a vocabulary $\mathcal{V}$ comprising chemical primitives (atoms, bonds, ring markers) and control symbols (\texttt{[BOS]}, \texttt{[EOS]}, \texttt{[PAD]}, \texttt{[MASK]}).

\textbf{Task formulation.}
We address two fundamental tasks:
\begin{enumerate}[leftmargin=*, nosep, label=(\roman*)]
    \item \emph{De novo generation}: Learn a generative model $p_\theta(\mathbf{x})$ that approximates the distribution of valid, drug-like, and synthetically accessible molecules.
    \item \emph{Target-specific molecular design}: Given a target protein, identify molecules $\mathbf{x}^*$ that maximize binding affinity while satisfying pharmacological constraints:
    \begin{equation}
        \begin{split}
            \mathbf{x}^* &= \operatorname*{argmin}_{\mathbf{x}} \mathrm{DS}(\mathbf{x}) \\
            \text{s.t.} &\quad \mathrm{QED}(\mathbf{x}) > \tau_{\text{QED}}, \; \mathrm{SA}(\mathbf{x}) < \tau_{\text{SA}},
        \end{split}
        \label{eq:sbdd_objective}
    \end{equation}

    where $\mathrm{DS}$ represents the docking score~\citep{alhossary2015qvina2} (lower indicates higher affinity), $\mathrm{QED}$ quantifies drug-likeness~\citep{bickerton2012qed}, and $\mathrm{SA}$ measures synthetic accessibility~\citep{ertl2009sa}.
\end{enumerate}

\subsection{Background: Generative Models for Sequences}
\label{sec:background}

\textbf{Autoregressive models} factorize the joint distribution as:
\begin{equation}
    p_\theta(\mathbf{x}) = \prod_{i=1}^{L} p_\theta(x_i \mid \mathbf{x}_{<i}).
    \label{eq:ar_factorization}
\end{equation}
While enabling efficient parallel training, this strictly causal factorization precludes bidirectional context. This restricts the modeling of local substructures, where chemical validity typically relies on the mutual dependencies of all constituent atoms rather than a linear sequence history~\citep{chemberta}.

\textbf{Discrete diffusion models}~\citep{austin2021structured} define a forward process $q(\mathbf{x}_t \mid \mathbf{x}_0)$ that progressively corrupts tokens, and learn a reverse denoising distribution $p_\theta(\mathbf{x}_0 \mid \mathbf{x}_t)$. Unlike autoregressive models, diffusion facilitates bidirectional context modeling, allowing the model to capture interactions among all tokens simultaneously. However, standard formulations typically operate on fixed-dimensional tensors~\citep{austin2021structured}, necessitating inefficient padding for molecular sequences. While variable-length extensions have been proposed~\citep{ssdlm, mdlm}, they often require complex length-modeling priors that complicate the simple termination mechanism inherent to autoregressive approaches.

\textbf{Block diffusion models}~\citep{bd3lm} unify these paradigms through a semi-autoregressive framework. This approach decomposes the sequence into tunable-granularity blocks modeled autoregressively, thereby retaining flexible termination while employing discrete diffusion to capture local bidirectional dependencies within each block:
\begin{equation}
    p_\theta(\mathbf{x}) = \prod_{b=1}^{B} p_\theta(\mathbf{x}^b \mid \mathbf{x}^{<b}).
    \label{eq:bd_model}
\end{equation}
In this formulation, each conditional $p_\theta(\mathbf{x}^b \mid \mathbf{x}^{<b})$ is parameterized as a denoising network. This design decouples global sequence generation from local structure modeling: the autoregressive prior manages sequence length and global coherence, while the block-wise diffusion employs bidirectional attention to capture the dense constraints of local chemical substructures.

\section{Method}
\label{sec:method}


\subsection{Soft-Fragment Representation}
\label{sec:softfragment}
We define \emph{soft-fragments} via a deterministic partitioning of a fixed-length molecular sequence.
Given a SMILES string $\mathbf{s}$, we first construct a dense sequence $\mathbf{x} = (x_1, \ldots, x_L)$ of uniform length $L$ by padding $\mathbf{s}$ with \texttt{[PAD]} tokens.
This sequence is then partitioned into $B = L/K$ contiguous blocks of fixed length $K$.
This approach constitutes a rule-free computational segmentation protocol (Figure~\ref{fig:paradigms}), where the $b$-th block is simply defined as:
\begin{equation}
\mathbf{x}^b := (x_{(b-1)K+1}, \ldots, x_{bK}),
\quad b \in \{1, \ldots, B\}.
\end{equation}
By strictly enforcing the global length $L$ to be a multiple of $K$, we ensure that every block possesses a uniform dimension without requiring conditional padding.
This formulation structurally decouples representation from generation: a fixed granularity $K_{\text{train}}$ is employed to capture intrinsic chemical syntax, while the sampling granularity $K_{\text{sample}}$ serves as a flexible hyperparameter to control search resolution during inference (detailed analysis in Appendix~\ref{app:effect_fragment}).

\subsection{SoftBD Architecture and Training}
\label{sec:blockdiffusion}

We implement SoftBD by instantiating Block Diffusion~\citep{bd3lm} over soft-fragments, following the factorization in \cref{eq:bd_model}.

\textbf{Block-wise attention masking.} We employ a hybrid attention mechanism~\citep{bd3lm} (Figure~\ref{fig:overview}(a)) on the concatenated noised ($\mathbf{x}_t$) and clean ($\mathbf{x}_0$) sequences. The applied mask $\mathcal{M}_{\text{full}} \in \{0, 1\}^{2L \times 2L}$ enforces inter-block causal dependencies while permitting intra-block bidirectional context. This structure is critical for soft-fragments: it enables the model to implicitly reconstruct chemical syntax and local bonds disrupted by arbitrary chunking, thereby eliminating the need for rule-based boundary preservation (details in Appendix~\ref{app:attention_mask}; empirical validation in Appendix~\ref{app:prefix_completion}).

\textbf{Training Objective.} We minimize the Block Diffusion Negative Evidence Lower Bound (NELBO)~\citep{bd3lm}. Leveraging the block-wise attention mask, we employ a vectorized training strategy to compute gradients for the entire molecule in a single forward pass.
Specifically, for each block $b$, we sample a time $t \sim \mathcal{U}[0, 1]$ and mask tokens with probability $1 - \alpha_t$.
The objective function aggregates the weighted cross-entropy losses over masked positions across all blocks (derivation in Appendix~\ref{app:training_objective}):
\begin{equation}
    \mathcal{L}_{\text{BD}}(\mathbf{x}; \theta) := \sum_{b=1}^{B} \mathbb{E}_{t, \mathbf{x}_t^b} \left[ -\frac{\alpha'_t}{1 - \alpha_t} \mathcal{L}_{\text{CE}}(\mathbf{x}^b, p_\theta(\cdot \mid \mathbf{x}_t^b, \mathbf{x}^{<b})) \right],
    \label{eq:bd_nelbo}
\end{equation}
where $\alpha_t$ denotes the noise schedule and $\alpha'_t$ its time derivative. Figure~\ref{fig:overview}(a) illustrates this pipeline.

\subsection{Adaptive Confidence Decoding for SoftBD}
\label{sec:efficient_inference}

SoftBD generates molecules semi-autoregressively (Figure~\ref{fig:overview}(b)): for each block $b=1,\ldots,B$, the model performs reverse diffusion conditioned on the cached history $\mathbf{x}^{<b}$. We implement \emph{Adaptive Confidence Decoding} (Algorithm~\ref{alg:inference}, Appendix~\ref{app:efficient_inference}), integrating three optimization mechanisms.
 
\textbf{First-Hitting Sampling.} Unlike standard masked diffusion which relies on a fixed, often redundant schedule $T$, we employ First-Hitting sampling~\citep{firsthitting} to analytically determine optimal transition times. For a sequence $i$ with $m_t^{(i)}$ masked tokens, the diffusion time updates via:
\begin{equation}
    t^{(i)}_{\text{next}} = t^{(i)}_{\text{curr}} \cdot (u^{(i)})^{1 / m^{(i)}_t}, \quad u^{(i)} \sim \mathcal{U}[0, 1].
\end{equation}
This mechanism dynamically aligns the step size with generative entropy: large $m_t$ triggers granular refinement steps, while small $m_t$ allows rapid progression, effectively eliminating computationally wasteful ``no-op'' iterations.

\textbf{Greedy Confidence Decoding.} To ensure high structural fidelity, we employ a deterministic strategy that prioritizes the resolution of high-confidence substructures. At each adaptive step, we identify the position-token pair $(j^\star, v^\star)$ that maximizes the local conditional probability across all masked indices $j \in \mathcal{M}_t$:
\begin{equation}
j^\star = \operatorname*{argmax}_{j \in \mathcal{M}_t} \left( \max_{v \in \mathcal{V}} p_\theta(x_j = v \mid \mathbf{x}_t^b, \mathbf{x}^{<b}) \right).
\label{eq:gcd}
\end{equation}
We deterministically unmask $x_{j^\star}$ to $v^\star$. This confidence-ordered revealing process prevents premature commitment to ambiguous tokens, ensuring that complex dependencies are resolved only when sufficient context is available. As shown in Table~\ref{tab:inference_ablation}, this strategy not only guarantees syntactic correctness but also substantially improves pharmacological quality.

\textbf{Batched Block-Wise Inference with Caching.} To maximize throughput, we parallelize generation by decoding mini-batches of molecules as independent Markov chains. Crucially, we employ a caching mechanism that freezes completed blocks as context, restricting active denoising computation solely to the current block. Coupled with a sliding attention window and an absorbing \texttt{[EOS]} termination token, this strategy amortizes the iterative refinement cost, achieving near-constant per-token latency.

\subsection{Gated MCTS for Target-Specific Search}
\label{sec:mcts}
We formulate target-specific molecular design as a Markov Decision Process (MDP) defined over the generative block space induced by SoftBD. A state $s_b$ represents a partial molecule comprising $b$ soft-fragments, and an action $a$ corresponds to the generation of the subsequent block $\mathbf{x}^{b+1} \sim p_\theta(\cdot \mid s_b)$. Unlike retrieval-based methods, this \emph{generative action space} eliminates reliance on rigid fragment libraries, enabling open-ended exploration with adaptive granularity controlled by $K_{\text{sample}}$. To effectively navigate this space under strict pharmacological constraints, we introduce a \textbf{Tunable Feasibility Gate} that decouples validity enforcement from binding affinity optimization. 
The search procedure follows four phases (Figure~\ref{fig:overview}(c); Algorithm~\ref{alg:mcts} in Appendix~\ref{app:mcts_details}).

\textbf{Selection.}
We guide tree traversal using a modified Upper Confidence Bound (UCT) that balances expectation and maximization. Each node $s$ tracks visit count $N(s)$, mean return $\bar{R}(s)$, and maximum return $R^{\max}(s)$. A child $s_j$ is selected by maximizing:
\begin{equation}
\mathrm{UCT}(s_j)
=
\lambda \bar{R}(s_j)
+
(1-\lambda) R^{\max}(s_j)
+
C\sqrt{\frac{\ln N(s)}{N(s_j)}},
\label{eq:uct}
\end{equation}
where $\lambda \in [0,1]$ modulates the trade-off between robust average performance and peak affinity discovery.
To efficiently allocate computational budget, we employ a Children-Adaptive widening strategy~\citep{tian2024imagination} that dynamically adjusts the branching factor based on value uncertainty. The node-specific expansion cap $C_{\mathrm{cap}}(s)$ is defined as:
\begin{equation}
\label{eq:adaptive_cap}
\begin{aligned}
C_{\mathrm{cap}}(s) &= \min\Bigl(C_{\max}, \max\bigl(C_{\min}, \lfloor \beta\, I(s)\rfloor\bigr)\Bigr), \\
I(s) &:= \max_{k}\bigl|\bar{R}(s_k)-\bar{R}(s)\bigr|,
\end{aligned}
\end{equation}
where $I(s)$ measures the dispersion of child values. This mechanism triggers wider expansion in regions with high variance (potential high rewards) while conserving resources in low-value or converged subspaces.

\textbf{Expansion.}
Upon reaching a leaf node $s_b$, we perform batched parallel exploration by sampling $M$ next-block candidates from $p_\theta(\cdot \mid s_b)$. After filtering duplicates, a unique candidate is instantiated as a new child node. This approach leverages SoftBD's batch inference capability to replace sequential rejection sampling with a single, high-throughput forward pass, efficiently discovering diverse transitions.

\textbf{Simulation with a Tunable Feasibility Gate.}
From the expanded child, we perform a rollout using the SoftBD prior to complete the molecule $\mathbf{x}$. To strictly decouple pharmacological compliance from affinity optimization, we implement a \emph{Tunable Feasibility Gate} parameterized by $(\tau_{\text{QED}},\tau_{\text{SA}})$. The reward function is defined as:
\begin{equation}
R(\mathbf{x}) =
\begin{cases}
-\mathrm{DS}(\mathbf{x}) & \text{if } \mathrm{QED}(\mathbf{x}) \ge \tau_{\text{QED}} \;\land\; \mathrm{SA}(\mathbf{x}) \le \tau_{\text{SA}},\\
R_{\mathrm{pen}} & \text{otherwise},
\end{cases}
\label{eq:feasibility_gate}
\end{equation}
where $\mathrm{DS}(\mathbf{x})$ denotes the Vina docking score and $R_{\mathrm{pen}}$ is a penalty for violation. This mechanism functions as a hierarchical computational sieve: by preemptively filtering candidates based on 2D properties, it ensures that computationally expensive 3D docking is reserved exclusively for pharmacologically viable structures, thereby maximizing the return on computational investment. 
We instantiate two configurations: \textbf{SoftMol} with default thresholds $(\tau_{\text{QED}}=0.5,\tau_{\text{SA}}=5.0)$, and \textbf{SoftMol (Unconstrained)} with $(\tau_{\text{QED}}=0,\tau_{\text{SA}}=\infty)$ to probe the affinity ceiling without pharmacological constraints.

\textbf{Backpropagation.}
Finally, the reward signal $R(\mathbf{x})$ is propagated recursively from the leaf to the root. We update the nodal statistics $\{N, \bar{R}, R^{\max}\}$ along the trajectory, progressively refining the tree policy to focus exploration on the optimal structural subspace defined by the feasibility gate and affinity landscape.

\begin{table*}[t]
    \centering
    \begin{minipage}[t]{0.7\textwidth}
        \caption{\textbf{\textit{De novo} molecule generation results.} The results are the means and standard deviations of 3 runs. $p$ and $\tau$ denote the nucleus sampling probability and softmax temperature, respectively. All SoftBD results use $K_{\text{sample}}=2$. The best results are highlighted in \textbf{bold}.}
        \vspace{1mm}
        \label{tab:denovo}
        \begin{small}
        \resizebox{\linewidth}{!}{%
        \begin{tabular}{lccccc}
            \toprule
            Method & Validity (\%) & Uniqueness (\%) & Quality (\%) & Docking-Filter (\%) & Diversity \\
            \midrule
            SAFE-GPT~\citep{safe2024} & $93.2 \pm 0.1$ & $\mathbf{100.0 \pm 0.0}$ & $54.4 \pm 0.6$ & $78.3 \pm 0.5$ & $0.879 \pm 0.000$ \\
            GenMol~\citep{genmol} & $99.9 \pm 0.1$ & $96.0 \pm 0.3$ & $85.2 \pm 0.4$ & $97.8 \pm 0.1$ & $0.817 \pm 0.000$ \\
            \midrule
            SoftBD ($p$=1.0, $\tau$=0.9) & $99.8 \pm 0.0$ & $\mathbf{100.0 \pm 0.0}$ & $87.1 \pm 0.2$ & $98.5 \pm 0.1$ & $0.871 \pm 0.000$ \\
            SoftBD ($p$=1.0, $\tau$=1.0) & $99.6 \pm 0.0$ & $\mathbf{100.0 \pm 0.0}$ & $84.7 \pm 0.2$ & $97.8 \pm 0.1$ & $0.878 \pm 0.000$ \\
            SoftBD ($p$=1.0, $\tau$=1.1) & $99.1 \pm 0.0$ & $\mathbf{100.0 \pm 0.0}$ & $81.7 \pm 0.3$ & $96.5 \pm 0.1$ & $0.883 \pm 0.000$ \\
            SoftBD ($p$=1.0, $\tau$=1.2) & $98.3 \pm 0.0$ & $\mathbf{100.0 \pm 0.0}$ & $77.7 \pm 0.3$ & $94.2 \pm 0.2$ & $0.888 \pm 0.000$ \\
            SoftBD ($p$=1.0, $\tau$=1.3) & $96.7 \pm 0.1$ & $\mathbf{100.0 \pm 0.0}$ & $72.9 \pm 0.3$ & $91.1 \pm 0.2$ & $\mathbf{0.893 \pm 0.000}$ \\
            \midrule
            SoftBD ($p$=0.95, $\tau$=0.9) & $\mathbf{100.0 \pm 0.0}$ & $98.4 \pm 0.1$ & $93.5 \pm 0.2$ & $99.8 \pm 0.0$ & $0.844 \pm 0.000$ \\
            SoftBD ($p$=0.95, $\tau$=1.0) & $\mathbf{100.0 \pm 0.0}$ & $99.4 \pm 0.1$ & $92.8 \pm 0.0$ & $99.7 \pm 0.0$ & $0.851 \pm 0.000$ \\
            SoftBD ($p$=0.95, $\tau$=1.1) & $\mathbf{100.0 \pm 0.0}$ & $99.6 \pm 0.1$ & $91.9 \pm 0.1$ & $99.6 \pm 0.0$ & $0.858 \pm 0.000$ \\
            SoftBD ($p$=0.95, $\tau$=1.2) & $99.9 \pm 0.0$ & $99.8 \pm 0.0$ & $90.8 \pm 0.1$ & $99.3 \pm 0.1$ & $0.867 \pm 0.000$ \\
            SoftBD ($p$=0.95, $\tau$=1.3) & $99.9 \pm 0.0$ & $99.8 \pm 0.1$ & $88.9 \pm 0.2$ & $98.9 \pm 0.1$ & $0.871 \pm 0.000$ \\
            \midrule
            SoftBD ($p$=0.9, $\tau$=0.9) & $\mathbf{100.0 \pm 0.0}$ & $90.0 \pm 0.2$ & $\mathbf{94.9 \pm 0.2}$ & $\mathbf{99.9 \pm 0.0}$ & $0.829 \pm 0.000$ \\
            SoftBD ($p$=0.9, $\tau$=1.0) & $\mathbf{100.0 \pm 0.0}$ & $96.0 \pm 0.1$ & $94.0 \pm 0.2$ & $99.8 \pm 0.0$ & $0.839 \pm 0.000$ \\
            SoftBD ($p$=0.9, $\tau$=1.1) & $\mathbf{100.0 \pm 0.0}$ & $98.0 \pm 0.1$ & $93.3 \pm 0.3$ & $99.8 \pm 0.0$ & $0.846 \pm 0.000$ \\
            SoftBD ($p$=0.9, $\tau$=1.2) & $\mathbf{100.0 \pm 0.0}$ & $99.1 \pm 0.1$ & $92.4 \pm 0.2$ & $99.7 \pm 0.0$ & $0.852 \pm 0.000$ \\
            SoftBD ($p$=0.9, $\tau$=1.3) & $\mathbf{100.0 \pm 0.0}$ & $99.3 \pm 0.1$ & $91.7 \pm 0.2$ & $99.6 \pm 0.0$ & $0.858 \pm 0.000$ \\
            \bottomrule
        \end{tabular}}
        \end{small}
    \end{minipage}
    \hfill
    \begin{minipage}[t]{0.29\textwidth}
        \centering
        \vspace{1mm}
        \includegraphics[width=\linewidth]{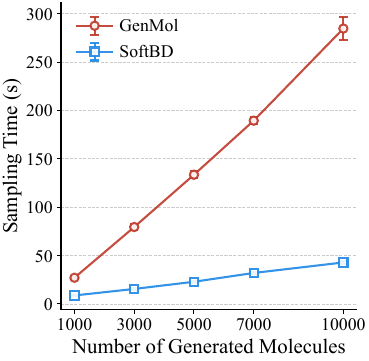}
        \vspace{-2mm}
        \makeatletter\def\@captype{figure}\makeatother
        \caption{Sampling time comparison between GenMol and SoftBD; SoftBD achieves $\approx 6.6\times$ shorter sampling time for 10k molecules.}
        \label{fig:sampling_time}
    \end{minipage}
    \vspace{-4mm}
\end{table*}

\section{Experiments}
\label{sec:experiments}
We systematically evaluate SoftMol across three key dimensions. First, we assess unsupervised generative performance and distribution matching against leading fragment-based baselines (\Cref{sec:denovo}). Second, we benchmark the full framework on target-specific molecular design tasks across diverse protein targets (\Cref{sec:structure_based}). Finally, we investigate the method's mechanistic underpinnings, including representation robustness and component-level ablations, in Appendix~\ref{app:effect_fragment} and Appendix~\ref{sec:experiments_ablation}.

\subsection{Datasets}
\label{sec:datasets}
We employ three large-scale datasets to evaluate general generative modeling and drug-likeness.

\textbf{SMILES (324M).} Following \citet{safe2024}, we aggregate a broad collection of molecules from ZINC~\citep{irwin2012zinc} and UniChem~\citep{chambers2013unichem}. We filter sequences exceeding 72 tokens (0.77\% rejection rate), yielding a final corpus of 324 million SMILES strings representing general chemical space.

\textbf{SAFE (322M).} To benchmark against rule-based fragmentation, we utilize the official SAFE dataset~\citep{safe2024}. From an initial pool of 326 million samples, we exclude sequences longer than 88 tokens (0.85\% removal) to ensure compatibility with fixed-window training, resulting in a corpus of 322 million SAFE strings.

\textbf{ZINC-Curated (427M).} To prioritize pharmaceutical relevance, we curate a high-quality subset of ZINC-22~\citep{sterling2015zinc} using a multi-stage filtration pipeline adapted from \citet{novomolgen}. The pipeline enforces physicochemical constraints, structural validity, and diversity-aware stratification (Appendix~\ref{app:dataset_details}). We impose a maximum SMILES length of $L=72$, which removes negligible data ($<0.001\%$) while retaining approximately 427 million drug-like molecules in SMILES format.

\subsection{\textit{De Novo} Generation}
\label{sec:denovo}

\textbf{Setup.} We train SoftBD (89M parameters) on ZINC-Curated, aligning model capacity with leading baselines to ensure fair comparison. We report performance across three dimensions: (1) \textbf{Standard Metrics}: Validity, Uniqueness, and Diversity, following MOSES~\citep{MOSES}; (2) \textbf{Quality}: the proportion of unique, valid molecules meeting strict drug-likeness criteria (QED $\ge 0.6$, SA $\le 4$)~\citep{genmol}; and (3) \textbf{Docking-Filter}: a pre-screening proxy for downstream docking viability defined by QED $> 0.5$ and SA $< 5$~\citep{lee2024GEAM}. For rigorous benchmarking, we evaluate all models on identical hardware, generating 10{,}000 samples per experiment across three independent seeds. Further details are provided in Appendix~\ref{app:de_novo_details}.

\textbf{Baselines.} We benchmark against two representative state-of-the-art fragment-based models. \textbf{SAFE-GPT}~\citep{safe2024} (87M parameters) employs an autoregressive model trained on 1.1 billion SAFE strings from ZINC and UniChem. \textbf{GenMol}~\citep{genmol} (89M parameters) is a discrete diffusion model trained on the same extensive corpus, distinguished by its high sampling efficiency. Crucially, SoftBD consumes less than 40\% of this training data (using only ZINC-Curated), thereby rigorously testing its data efficiency relative to these large-scale pre-trained baselines.

\textbf{Results.} Table~\ref{tab:denovo} highlights SoftBD's robust generative performance. Despite training on a significantly smaller corpus, SoftBD matches or exceeds the large-scale baselines SAFE-GPT and GenMol on strict drug-likeness metrics (Quality and Docking-Filter) while preserving superior structural diversity. Notably, the model achieves 100\% validity across most configuration settings, effectively refuting the assumption that explicit rule encoding is requisite for syntactic robustness in molecule generation. This robustness is further corroborated in Appendix~\ref{app:prefix_completion}, where we demonstrate that SoftBD successfully repairs syntactically broken prefixes (e.g., unclosed rings) with near-perfect accuracy, confirming that the model learns intrinsic chemical syntax rather than relying on rigid tokenization rules. We also observe a controllable trade-off between quality and exploration: lower temperatures ($\tau$) and nucleus probabilities ($p$) shift the distribution towards high-fidelity drug-like candidates, while relaxed constraints facilitate broad exploration. Computationally, SoftBD delivers a $\approx$6.6$\times$ acceleration over the discrete diffusion baseline GenMol (\Cref{fig:sampling_time}). Collectively, these gains underscore the synergy between the Block-Diffusion architecture, Adaptive Confidence Decoding, and the high-quality ZINC-Curated training corpus.

\begin{table*}[t]
    \centering
    \caption{\textbf{Novel top-hit 5\% docking score (kcal/mol) results.} The results are the means and the standard deviations of 3 runs. Lower is better; the best results are highlighted in \textbf{bold} and the second best are \underline{underlined}.}
    \vspace{-1mm}
    \label{tab:docking}
    \begin{small}
	    \resizebox{\textwidth}{!}{%
	    \begin{tabular}{lccccc}
	        \toprule
	        \multirow{2}{*}{Method} & \multicolumn{5}{c}{Target protein} \\
	        \cmidrule(lr){2-6}
	         & parp1 & fa7 & 5ht1b & braf & jak2 \\
        \midrule
        REINVENT~\citep{reinvent} & $-8.702 \pm 0.523$ & $-7.205 \pm 0.264$ & $-8.770 \pm 0.316$ & $-8.392 \pm 0.400$ & $-8.165 \pm 0.277$ \\
        JTVAE~\citep{jin2018jtvae} & $-9.482 \pm 0.132$ & $-7.683 \pm 0.048$ & $-9.382 \pm 0.332$ & $-9.079 \pm 0.069$ & $-8.885 \pm 0.026$ \\
        GA+D~\citep{nigam2019} & $-8.365 \pm 0.201$ & $-6.539 \pm 0.297$ & $-8.567 \pm 0.177$ & $-9.371 \pm 0.728$ & $-8.610 \pm 0.104$ \\
        Graph-GA~\citep{graphga} & $-10.949 \pm 0.532$ & $-7.365 \pm 0.326$ & $-10.422 \pm 0.670$ & $-10.789 \pm 0.341$ & $-10.167 \pm 0.576$ \\
        MORLD~\citep{morld} & $-7.532 \pm 0.260$ & $-6.263 \pm 0.165$ & $-7.869 \pm 0.650$ & $-8.040 \pm 0.337$ & $-7.816 \pm 0.133$ \\
        HierVAE~\citep{jin2020hiervae} & $-9.487 \pm 0.278$ & $-6.812 \pm 0.274$ & $-8.081 \pm 0.252$ & $-8.978 \pm 0.525$ & $-8.285 \pm 0.370$ \\

        GEGL~\citep{gegl} & $-9.329 \pm 0.170$ & $-7.470 \pm 0.013$ & $-9.086 \pm 0.067$ & $-9.073 \pm 0.047$ & $-8.601 \pm 0.038$ \\
        RationaleRL~\citep{rationaleRL} & $-10.663 \pm 0.086$ & $-8.129 \pm 0.048$ & $-9.005 \pm 0.155$ & \textit{No hit found} & $-9.398 \pm 0.076$ \\
        FREED~\citep{freed} & $-10.579 \pm 0.104$ & $-8.378 \pm 0.044$ & $-10.714 \pm 0.183$ & $-10.561 \pm 0.080$ & $-9.735 \pm 0.022$ \\
        MARS~\citep{mars} & $-9.716 \pm 0.082$ & $-7.839 \pm 0.018$ & $-9.804 \pm 0.073$ & $-9.569 \pm 0.078$ & $-9.150 \pm 0.114$ \\
        PS-VAE~\citep{ps_vae} & $-9.978 \pm 0.091$ & $-8.028 \pm 0.050$ & $-9.887 \pm 0.115$ & $-9.637 \pm 0.049$ & $-9.464 \pm 0.129$ \\
        MOOD~\citep{mood} & $-10.865 \pm 0.113$ & $-8.160 \pm 0.071$ & $-11.145 \pm 0.042$ & $-11.063 \pm 0.034$ & $-10.147 \pm 0.060$ \\
        RetMol~\citep{retmol} & $-8.590 \pm 0.475$ & $-5.448 \pm 0.688$ & $-6.980 \pm 0.740$ & $-8.811 \pm 0.574$ & $-7.133 \pm 0.242$ \\
        Genetic GFN~\citep{genetic_gfn} & $-9.227 \pm 0.644$ & $-7.288 \pm 0.433$ & $-8.973 \pm 0.804$ & $-8.719 \pm 0.190$ & $-8.539 \pm 0.592$ \\
        GEAM~\citep{lee2024GEAM} & $-12.891 \pm 0.158$ & $-9.890 \pm 0.116$ & $-12.374 \pm 0.036$ & $-12.342 \pm 0.095$ & $-11.816 \pm 0.067$ \\
        $f$-RAG~\citep{lee2024frag} & $-12.945 \pm 0.053$ & $-9.899 \pm 0.205$ & $-12.670 \pm 0.144$ & $-12.390 \pm 0.046$ & $-11.842 \pm 0.316$ \\
        GenMol~\citep{genmol} & $-11.773 \pm 0.332$ & $-8.967 \pm 0.289$ & $-11.914 \pm 0.183$ & $-11.394 \pm 0.203$ & $-10.417 \pm 0.229$ \\
        \midrule
        \textbf{SoftMol} & \underline{$-13.977 \pm 0.044$} & \underline{$-11.031 \pm 0.045$} & \underline{$-13.733 \pm 0.031$} & \underline{$-13.252 \pm 0.018$} & \underline{$-12.580 \pm 0.007$} \\
        \textbf{SoftMol (Unconstrained)} & $\mathbf{-14.168 \pm 0.072}$ & $\mathbf{-11.235 \pm 0.047}$ & $\mathbf{-13.903 \pm 0.030}$ & $\mathbf{-13.430 \pm 0.032}$ & $\mathbf{-12.807 \pm 0.039}$ \\
        \bottomrule 
    \end{tabular}}
    \end{small}
\end{table*}

\begin{figure*}[t]
    \vspace{-2mm}
    \centering
    \includegraphics[width=\textwidth]{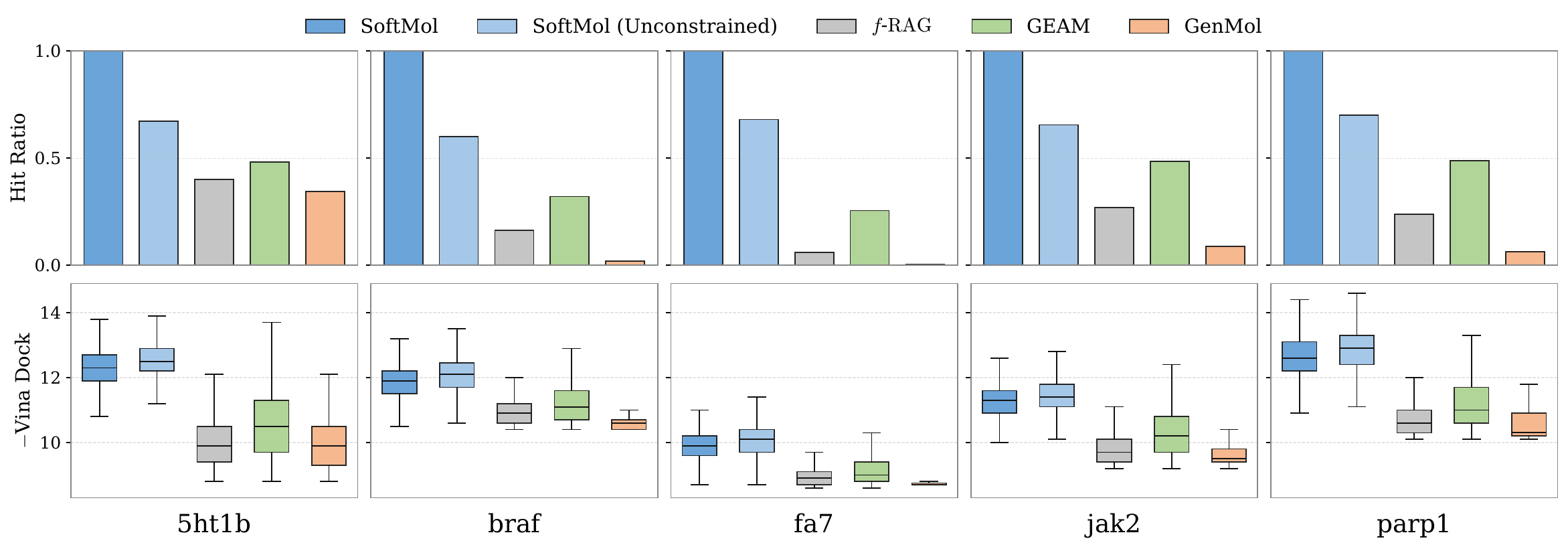}
    \vspace{-4mm}
    \caption{\textbf{Hit Ratio and DS Distributions.} \textbf{Top:} Hit Ratio, defined as the proportion of unique generated molecules simultaneously satisfying drug-likeness (QED $> 0.5$, SA $< 5.0$) and binding affinity (DS $<$ median active) criteria. \textbf{Bottom:} Distribution of negative docking scores ($-$DS) for the identified hits satisfying all three criteria. Higher values indicate stronger affinity.}
    \label{fig:combined_hit_rates}
    \vspace{-4mm}
\end{figure*}

\begin{table*}[t]
    \centering
    \begin{minipage}[t]{0.7\textwidth}
        \caption{\textbf{\#Circles of generated hit molecules.} The \#Circles threshold is set to 0.75. The results are the means and standard deviations of 3 runs. The best results are highlighted in \textbf{bold} and the second best are \underline{underlined}.}
        \label{tab:circles}
        \begin{small}
        \resizebox{\linewidth}{!}{%
        \begin{tabular}{lccccc}
            \toprule
            \multirow{2}{*}{Method} & \multicolumn{5}{c}{Target protein} \\
            \cmidrule(lr){2-6}
             & parp1 & fa7 & 5ht1b & braf & jak2 \\
            \midrule
            REINVENT~\citep{reinvent} & $44.2 \pm 15.5$ & $23.2 \pm 6.6$ & $138.8 \pm 19.4$ & $18.0 \pm 2.1$ & $59.6 \pm 8.1$ \\
            MORLD~\citep{morld} & $1.4 \pm 1.5$ & $0.2 \pm 0.4$ & $22.2 \pm 16.1$ & $1.4 \pm 1.2$ & $6.6 \pm 3.7$ \\
            HierVAE~\citep{jin2020hiervae} & $4.8 \pm 1.6$ & $0.8 \pm 0.7$ & $5.8 \pm 1.0$ & $3.6 \pm 1.4$ & $4.8 \pm 0.7$ \\
            RationaleRL~\citep{rationaleRL} & $61.3 \pm 1.2$ & $2.0 \pm 0.0$ & $\mathbf{312.7 \pm 6.3}$ & $1.0 \pm 0.0$ & $199.3 \pm 7.1$ \\
            FREED~\citep{freed} & $34.8 \pm 4.9$ & $21.2 \pm 4.0$ & $88.2 \pm 13.4$ & $34.4 \pm 8.2$ & $59.6 \pm 8.2$ \\
            PS-VAE~\citep{ps_vae} & $38.0 \pm 6.4$ & $18.0 \pm 5.9$ & $180.7 \pm 11.6$ & $16.0 \pm 0.8$ & $83.7 \pm 11.9$ \\
            MOOD~\citep{mood} & $86.4 \pm 11.2$ & $19.2 \pm 4.0$ & $144.4 \pm 15.1$ & $50.8 \pm 3.8$ & $81.8 \pm 5.7$ \\
            GEAM~\citep{lee2024GEAM} & $123.0 \pm 7.8$ & $79.0 \pm 9.2$ & $144.3 \pm 8.6$ & $84.7 \pm 8.6$ & $118.3 \pm 0.9$ \\
            $f$-RAG~\citep{lee2024frag} & $54.7 \pm 8.1$ & $31.7 \pm 3.1$ & $64.3 \pm 18.2$ & $36.0 \pm 10.6$ & $54.7 \pm 7.4$ \\
            GenMol~\citep{genmol} & $3.0 \pm 0.0$ & $2.3 \pm 0.6$ & $20.3 \pm 2.1$ & $6.3 \pm 1.5$ & $2.3 \pm 0.6$ \\
            \midrule
            \textbf{SoftMol} & $\mathbf{215.0 \pm 6.2}$ & $\mathbf{254.3 \pm 10.2}$ & \underline{$246.3 \pm 21.1$} & $\mathbf{291.7 \pm 17.0}$ & $\mathbf{330.0 \pm 18.5}$ \\
            \textbf{SoftMol (Unconstrained)} & \underline{$160.3 \pm 3.5$} & \underline{$213.0 \pm 4.6$} & $186.7 \pm 10.3$ & \underline{$217.7 \pm 13.6$} & \underline{$264.0 \pm 3.0$} \\
            \bottomrule
        \end{tabular}}
        \end{small}
    \end{minipage}
    \hfill
    \begin{minipage}[t]{0.28\textwidth}
        \centering
        \vspace{1mm}
        \includegraphics[width=\linewidth]{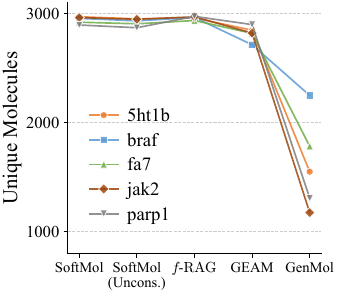}
        \makeatletter\def\@captype{figure}\makeatother
        \caption{Number of unique molecules per 3,000 attempts. SoftMol maintains high uniqueness.}
        \label{fig:uniqueness}
    \end{minipage}
    \vspace{-4mm}
\end{table*}

\subsection{Target-specific molecular design}
\label{sec:structure_based}

\textbf{Setup.} Following the protocol of \citet{lee2024frag}, we benchmark SoftMol on protein-targeted generation tasks across five targets: parp1, fa7, 5ht1b, braf, and jak2. For each target, we generate 3,000 molecules across three independent runs.
The primary evaluation metric is the \textbf{Novel Top-hit 5\% Score}, defined as the mean docking score (DS) of the top 5\% unique and novel generated hits.
In contrast to methods such as $f$-RAG~\citep{lee2024frag} and GEAM~\citep{lee2024GEAM}, which are trained or fine-tuned on relatively small-scale datasets like ZINC250K, SoftBD is pretrained on a large-scale molecular corpus without target-specific adaptation.
Accordingly, we define a \textbf{Novel Hit} with three strict criteria to emphasize both affinity and drug-likeness: (1)~DS $<$ median DS of known actives; (2)~QED $> 0.5$; and (3)~SA $< 5.0$.
We additionally report:
\vspace{-1mm}
\begin{itemize}[leftmargin=*, nosep]
    \item \textbf{Uniqueness:} The number of unique valid molecules generated out of 3,000 attempts, indicating the absence of mode collapse.
    \item \textbf{Hit Ratio:} The proportion of unique generated molecules that qualify as novel hits, measuring generative success rate.
    \item \textbf{\#Circles:} The number of distinct structural clusters among novel hits~\citep{xie2023circles}, measuring diversity and coverage of the explored chemical space.
\end{itemize}
\vspace{-1mm}
We report results for both \textbf{SoftMol} and \textbf{SoftMol (Unconstrained)}, as defined in \Cref{sec:mcts}. Implementation details are provided in Appendix~\ref{app:structure_based_details}.

\textbf{Baselines.} We benchmark against 17 state-of-the-art methods spanning four categories. (1) \textit{Fragment-Based Methods}: \textbf{JT-VAE} \citep{jin2018jtvae}, \textbf{HierVAE} \citep{jin2020hiervae}, \textbf{MARS} \citep{mars}, \textbf{RationaleRL} \citep{rationaleRL}, \textbf{FREED} \citep{freed}, \textbf{PS-VAE} \citep{ps_vae}, \textbf{GEAM} \citep{lee2024GEAM}, \textbf{$f$-RAG} \citep{lee2024frag}, and \textbf{RetMol} \citep{retmol}. (2) \textit{Evolutionary Algorithms}: \textbf{Graph-GA} \citep{graphga}, \textbf{GEGL} \citep{gegl}, \textbf{Genetic-GFN} \citep{genetic_gfn}, and \textbf{GA+D} \citep{nigam2019}. (3) \textit{Reinforcement Learning}: \textbf{REINVENT} \citep{reinvent} and \textbf{MORLD} \citep{morld}. (4) \textit{Diffusion Models}: \textbf{MOOD} \citep{mood} and \textbf{GenMol} \citep{genmol}. 
Binding affinity results in \Cref{tab:docking} are reported from \citet{lee2024frag}, excluding GenMol. Similarly, diversity metrics in \Cref{tab:circles} are sourced from \citet{lee2024GEAM}, with the exception of GenMol. We fully reproduce GenMol and all results in \Cref{fig:combined_hit_rates,fig:uniqueness} using their official implementations.

\textbf{Results.}    \emph{Binding Affinity.}
As evidenced in \Cref{tab:docking}, a new state-of-the-art is established by SoftMol. Superior binding affinity is consistently achieved across all five protein targets compared to 17 baselines, despite the enforcement of strict pharmacological constraints. Notably, a substantial performance margin is observed over leading fragment-based methods, such as $f$-RAG and GEAM. This effectiveness is attributed to the generative soft-fragment action space, which facilitates the exploration of novel chemotypes beyond the rigid, predefined vocabularies inherent to prior approaches. Furthermore, the theoretical affinity ceiling is quantified by the unconstrained variant, which yields the lowest docking scores across all evaluated models.

\emph{Hit Ratio and DS Distributions.}
\Cref{fig:combined_hit_rates} evaluates generation efficiency and hit quality.
SoftMol consistently achieves a \textbf{100\% Hit Ratio} across all targets, confirming that the feasibility gate effectively constrains the search to the drug-like manifold without impeding the discovery of high-affinity binders.
Conversely, the lower hit ratios observed in baselines (e.g., GenMol, $f$-RAG) indicate a failure to balance affinity optimization with physicochemical validity.
Notably, SoftMol (Unconstrained) outperforms most baselines in hit ratio even without explicit gating, suggesting that the pharmaceutically relevant chemical space is intrinsically captured by the pretrained prior.
Furthermore, the docking score distributions (bottom) demonstrate that SoftMol closely approximates the affinity ceiling established by the unconstrained variant, implying that strict pharmacological compliance incurs negligible performance cost.

\emph{Diversity and Exploration.}
Finally, we investigate whether the improvements in binding affinity come at the cost of generative diversity, specifically regarding mode collapse.
As shown in \Cref{fig:uniqueness}, both SoftMol configurations consistently generate nearly 3,000 unique candidates per 3,000 attempts. This confirms that the MCTS policy maintains robust exploration, avoiding the local optima traps that lead to severe redundancy in baselines like GenMol.
Beyond uniqueness, structural diversity is quantified using the \#Circles metric (\Cref{tab:circles}). SoftMol achieves 2--3$\times$ higher diversity than the leading baseline (GEAM), validating that the generative soft-fragment space enables the construction of novel chemotypes that transcend the rigid vocabularies of retrieval-based methods.
Interestingly, SoftMol (Unconstrained) exhibits slightly reduced diversity compared to the constrained variant. We attribute this to aggressive exploitation: without feasibility barriers, the search converges more intensely toward deep high-affinity basins. Nevertheless, even in this unconstrained regime, SoftMol significantly outperforms all baselines, demonstrating that its exploratory power is intrinsic to the representation and resilient to strong optimization pressures.

\begin{figure*}[t]
    \centering
    \includegraphics[width=\textwidth]{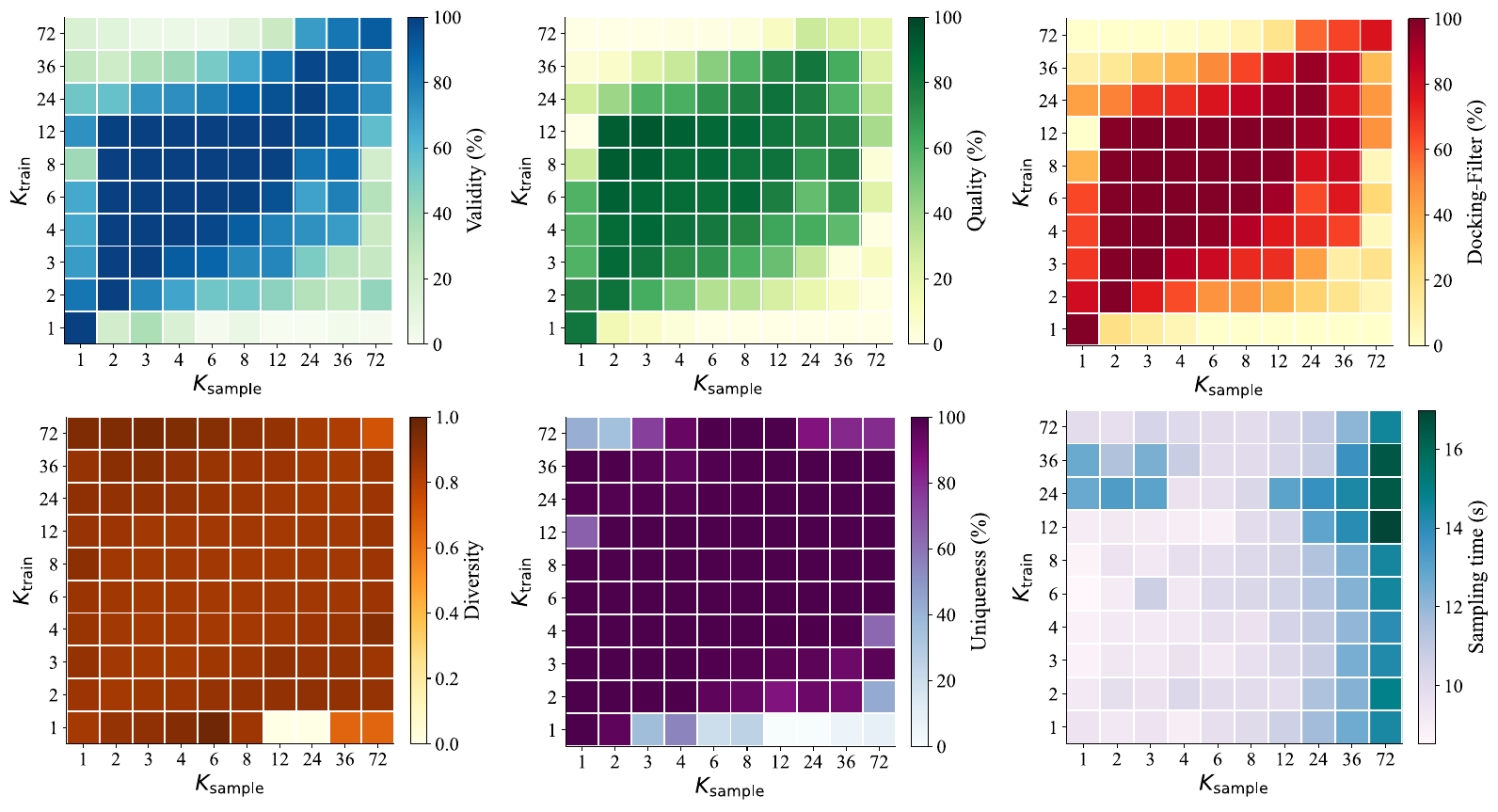}
    \caption{\textbf{Effect of Soft-Fragment Length.} Heatmaps display Validity, Quality, Docking-Filter, Diversity, Uniqueness, and Sampling Time across the full $K_{\text{train}} \times K_{\text{sample}}$ grid.}
    \label{fig:robustness_heatmap}
    \vspace{-5mm}
\end{figure*}

\section{Analysis of Soft-Fragment Length}
\label{app:effect_fragment}
We investigate how the soft-fragment length governs the trade-off between global autoregressive modeling and local bidirectional diffusion.
Unlike prior methods that couple training and generation granularities (e.g., token-level AR or fixed-fragment diffusion), SoftMol decouples them: $K_{\text{train}}$ acts as the representation granularity for learning chemical syntax, while $K_{\text{sample}}$ serves as a flexible inference knob controlling search step size.
To characterize this interaction, we systematically train models with varying representation granularities $K_{\text{train}} \in \{1, 2, 3, 4, 6, 8, 12, 24, 36, 72\}$ and evaluate each across the same set of sampling granularities $K_{\text{sample}}$.
The resulting performance landscape, visualized in Figure~\ref{fig:robustness_heatmap}, reveals how SoftBD balances representation learning with decoding dynamics.Experimental details and further efficiency analysis are provided in Appendix~\ref{app:soft_fragment_length_details}.

\textbf{Robustness of Pharmacological Quality.}
Figure~\ref{fig:robustness_heatmap} (top row) identifies a broad ``High-Performance Plateau'' for $K_{\text{train}} \in [4, 12]$, where the model consistently achieves near-perfect Validity ($>99.9\%$) and high Quality ($>80\%$) across a wide range of sampling granularities ($K_{\text{sample}} \in [2, 12]$).
Similarly, the Docking-Filter rate follows this same high-performance plateau, confirming that drug-likeness capabilities are preserved whenever representation and sampling steps are aligned.
This plateau highlights the stability of the soft-fragment representation.
Performance degrades only at the pathological extremes: at $K_{\text{train}}=1$, the lack of local bidirectional context leads to brittle syntax handling (Validity drops); at $K_{\text{train}} \ge 24$, the diffusion decoder becomes under-conditioned, failing to reconstruct complex substructures (Quality drops).
Crucially, within the stable region, the model's grasp of chemical syntax is not tied to a specific tokenization boundary. 

\begin{table*}[t]
    \centering
    \begin{minipage}[t]{0.49\textwidth}
        \caption{\textbf{Ablation of Adaptive Confidence Decoding.} The results represent the means and standard deviations across 3 independent runs, with 1,000 molecules generated per run. All results use $K_{\text{sample}}=8$. The best results are highlighted in \textbf{bold}.}
        \label{tab:inference_ablation}
        \centering
        \small
        \resizebox{\linewidth}{!}{
        \begin{tabular}{ccccccc}
        \toprule
        \multicolumn{3}{c}{Components} & \multicolumn{4}{c}{Metrics} \\
        \cmidrule(lr){1-3}\cmidrule(lr){4-7}
        FH & GCD & Batch & Validity (\%) & Quality (\%) & Diversity & Time (s) \\
        \midrule
        \textsc{off} & \textsc{off} & \textsc{off} & 88.6 $\pm$ 1.0 & 55.4 $\pm$ 1.5 & \textbf{0.855 $\pm$ 0.001} & 1308.4 $\pm$ 7.4 \\
        \textsc{on}  & \textsc{off} & \textsc{off} & 90.1 $\pm$ 0.5 & 55.2 $\pm$ 1.8 & 0.853 $\pm$ 0.000 & 512.5 $\pm$ 14.7 \\
        \textsc{on}  & \textsc{on}  & \textsc{off} & \textbf{100.0 $\pm$ 0.0} & 80.4 $\pm$ 0.9 & 0.845 $\pm$ 0.001 & 471.6 $\pm$ 32.7 \\
        \textsc{on}  & \textsc{off} & \textsc{on}  & 90.6 $\pm$ 1.0 & 55.7 $\pm$ 0.7 & 0.854 $\pm$ 0.000 & \textbf{9.6 $\pm$ 0.5} \\
        \textsc{on}  & \textsc{on}  & \textsc{on}  & \textbf{100.0 $\pm$ 0.0} & \textbf{81.9 $\pm$ 2.0} & 0.845 $\pm$ 0.001 & 10.3 $\pm$ 0.9 \\
        \bottomrule
        \end{tabular}
        }
    \end{minipage}
    \hfill
    \begin{minipage}[t]{0.49\textwidth}
        \caption{\textbf{Ablation of pretraining corpus and model scale.} The results represent the means and standard deviations across 3 independent runs, with 1,000 molecules generated per run. All results use $K_{\text{sample}}=8$. The best results are highlighted in \textbf{bold}.}
        \label{tab:corpus_scaling}
        \centering
        \small
        \resizebox{\linewidth}{!}{
        \begin{tabular}{lccccc}
        \toprule
        Dataset & Params & Validity (\%) & Quality (\%) & Diversity & Time (s) \\
        \midrule
        \multirow{2}{*}{SMILES}
         & 55M & \textbf{100.0} $\pm$ 0.0 & 61.4 $\pm$ 1.7 & 0.854 $\pm$ 0.001 & \textbf{5.357 $\pm$ 0.660} \\
         & 72M &  99.9 $\pm$ 0.0 & 61.4 $\pm$ 2.1 & 0.852 $\pm$ 0.000 & 8.167 $\pm$ 0.339 \\
        \midrule
        SAFE & 74M & 91.9 $\pm$ 0.0 & 59.0 $\pm$ 0.0 & \textbf{0.870 $\pm$ 0.002} & 9.487 $\pm$ 0.434 \\
        \midrule
        \multirow{5}{*}{ZINC-Curated}
         & 55M  &  99.9 $\pm$ 0.1 & 81.0 $\pm$ 0.7 & 0.845 $\pm$ 0.001 & 5.975 $\pm$ 0.565 \\
         & 72M  &  99.9 $\pm$ 0.1 & 80.6 $\pm$ 0.3 & 0.845 $\pm$ 0.001 & 7.865 $\pm$ 0.586 \\
         & 89M  & \textbf{100.0 $\pm$ 0.0} & \textbf{81.9 $\pm$ 2.0} & 0.845 $\pm$ 0.001 & 8.896 $\pm$ 0.377 \\
         & 116M & \textbf{100.0 $\pm$ 0.0} & 80.7 $\pm$ 0.4 & 0.844 $\pm$ 0.000 & 10.976 $\pm$ 0.667 \\
         & 624M & \textbf{100.0 $\pm$ 0.0} & 81.0 $\pm$ 0.4 & 0.846 $\pm$ 0.001 & 33.965 $\pm$ 0.928 \\
        \bottomrule
        \end{tabular}
        }
    \end{minipage}
\end{table*}

\textbf{Diversity and Computational Efficiency.}
Regarding exploration capability, the Diversity metric (bottom-left) remains consistently high across the entire grid, indicating that the soft-fragment representation robustly covers chemical space without collapsing into repetitive modes.
Complementing this, Uniqueness (bottom-middle) generally mirrors Quality, remaining above 99\% across the stability region and dropping only due to mismatched granularities.
However, Sampling Time (bottom-right) reveals a critical practical constraint: while latency remains stable at $\approx 9$--10\,s for blocks $K_{\text{sample}} \le 12$, it climbs to 13--17\,s for extremely large blocks ($K_{\text{sample}} \ge 24$).
This occurs because the diffusion attention mechanism scales quadratically with block length, outweighing the benefits of fewer autoregressive steps.
Although $K_{\text{sample}}=1$ offers low latency, it severely degrades pharmacological Quality.
Taken together, the range $K_{\text{sample}} \in [2, 12]$ offers the optimal trade-off between quality and efficiency.

\textbf{Task-Adaptive Granularity Selection.}
The decoupled nature of SoftMol enables inference-time adaptation without retraining.
Leveraging the robustness observed in the plateau, we select $K_{\text{train}}=8$ as our canonical configuration and vary $K_{\text{sample}}$ to maximize performance for specific downstream tasks.
For \emph{de novo} generation, we use a finer granularity ($K_{\text{sample}}=2$) to maximize local precision and Quality.
Conversely, for target-specific optimization via MCTS, we employ a coarser granularity ($K_{\text{sample}}=8$) to expand larger valid substructures per step, effectively reducing the search depth and computational cost.
This flexibility---dictated by the task rather than the training scheme---is a unique advantage of the computation-centric soft-fragment paradigm.

\section{Ablation Studies}
\label{sec:experiments_ablation}

To rigorously validate the design of SoftMol, we dissect the framework into two core dimensions: \emph{Generative Foundations} and \emph{Search \& Inference Dynamics}.
First, we quantify the efficiency gains from our decoding strategies (\Cref{tab:inference_ablation}) and isolate the impact of data quality and representation efficiency (\Cref{tab:corpus_scaling}).
Second, we analyze the scaling behavior of the model and the interplay between granularity and search budget in Gated MCTS (\Cref{fig:ablation_granularity}).
Detailed experimental protocols are provided in Appendix~\ref{app:ablation_details}.

\textbf{Ablation of Adaptive Confidence Decoding.}
Table~\ref{tab:inference_ablation} quantifies the hierarchical contribution of each inference component.
First, \emph{First-Hitting Sampling (FH)} yields a $2.6\times$ speedup by dynamically eliminating redundant denoising steps.
Second, \emph{Greedy Confidence Decoding (GCD)} functions as a critical quality enhancement, bridging the validity gap ($88.6\% \to 100.0\%$) and substantially improving pharmacological quality by $26\%$ ($55.4\% \to 81.9\%$).
Third, \emph{Batched Inference (Batch)} unlocks massive parallelism, amplifying throughput by $46\times$ compared to non-batched execution.
Collectively, this integrated strategy achieves a $130\times$ acceleration while guaranteeing perfect structural validity, facilitating efficient large-scale screening.

\textbf{Ablation of Pretraining Corpus.} 
Modern NLP has established that high-quality generation requires both pretraining and post-training alignment~\citep{achiam2023gpt4}. We demonstrate that a carefully curated pretraining corpus alone can substantially improve pharmacological quality without requiring additional post-training techniques such as DPO or reinforcement learning.
To disentangle architectural inductive biases from data distribution effects, we compare SoftBD models trained on raw SMILES versus our curated ZINC-Curated (\Cref{tab:corpus_scaling}). We observe that in our experiments, validity remains high across both corpora, suggesting that the soft-fragment Block-Diffusion architecture effectively maintains structural correctness. In contrast, the pharmacological profile faithfully mirrors the training distribution: transitioning to the curated dataset yields a substantial gain in Quality with negligible degradation in Diversity. This demonstrates that SoftBD acts as a high-fidelity generative model that effectively decouples syntax learning from property optimization, while simultaneously validating our dataset's contribution in shifting the generative Pareto frontier by filtering non-drug-like noise without sacrificing chemical space diversity.

\textbf{Ablation of Representation Efficiency.}
We additionally evaluate the SAFE representation~\citep{safe2024} under our Block-Diffusion architecture. Despite SAFE's design goal of enforcing chemical validity through rule-based fragmentation, SAFE dataset yields lower Validity (91.9\%) and Quality (59.0\%) compared to both raw SMILES and ZINC-Curated (\Cref{tab:corpus_scaling}). We attribute this to two factors: (1) SAFE's auxiliary connectivity tokens and variable-length fragments inflate sequence complexity, leading to higher token consumption per molecule; and (2) the rigid fragmentation heuristics conflict with our architecture's inductive bias, which learns chemical syntax directly from fixed-length soft-fragments. These results suggest that explicit rule-based representations are not only unnecessary for validity but may actively hinder generation quality when combined with architectures designed for computation-centric representations.

\textbf{Ablation of Model Scaling.}
Given that the curated ZINC-Curated dataset yields superior pharmacological quality, we adopt it as the standard corpus to examine the scaling properties of SoftMol.
We train models ranging from 55M to 624M parameters and observe a rapid saturation of generation quality: a compact 55M model already attains strong performance in Quality and Docking-Filter, with only marginal fluctuations when scaling to 624M (\Cref{tab:corpus_scaling}).
Sampling latency nevertheless grows roughly linearly with parameter count.
These findings demonstrate that the superior performance of SoftMol over fragment-based baselines is driven by the efficiency of the representation and architecture rather than brute-force model scaling.
Consequently, the 89M model utilized in our main experiments represents a Pareto-optimal choice, delivering saturated quality with a highly favorable computational footprint.

\begin{figure*}[t]
    \centering
    \includegraphics[width=\textwidth]{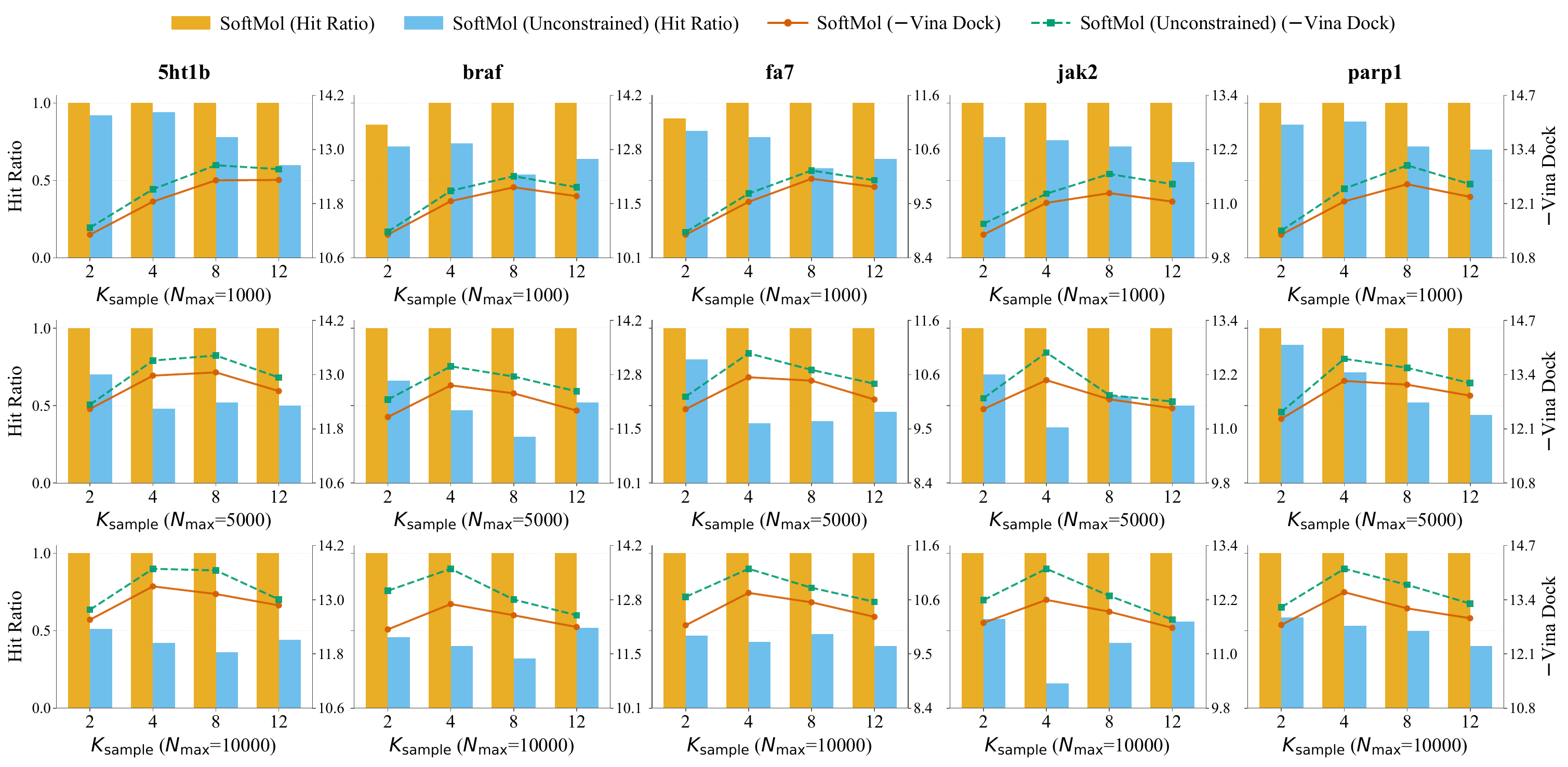}
    \caption{\textbf{Ablation on Granularity and Search Budget.} Hit Ratio and docking scores for varying $K_{\text{sample}}$ across different search budgets $N_{\max} \in \{1000, 5000, 10000\}$. Results are averaged across 5 targets from 50 independent runs.}
    \label{fig:ablation_granularity}
\end{figure*}

\begin{figure}[t]
    \centering
    \includegraphics[width=1.0\linewidth]{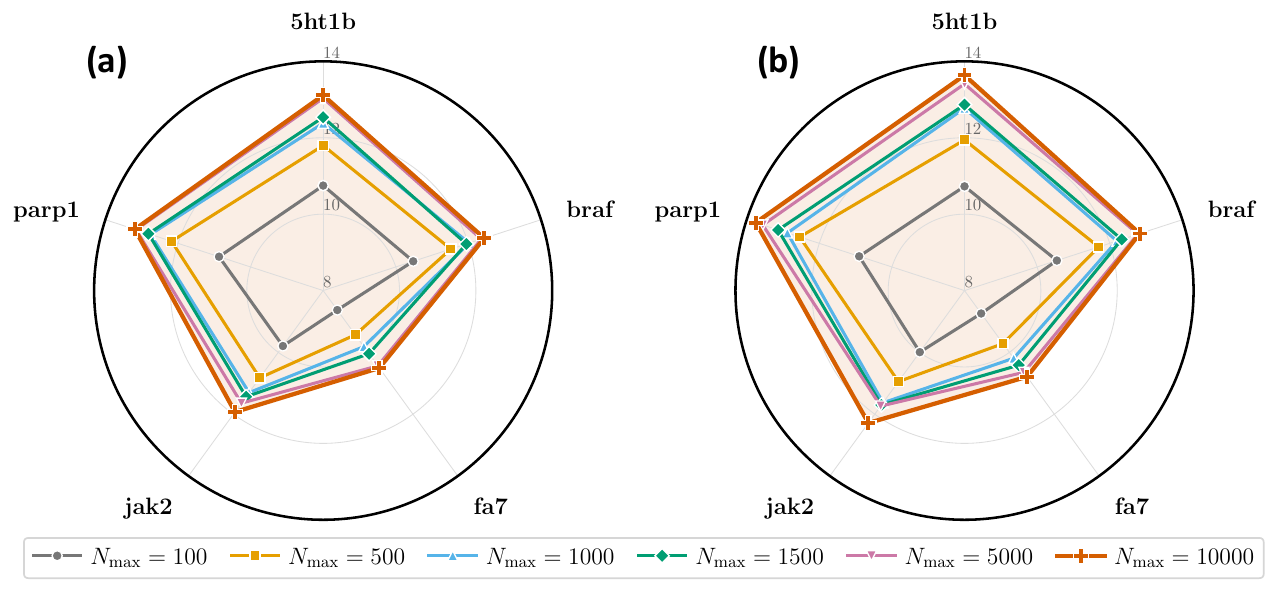}
    \caption{\textbf{Search Budget Scaling.} \textbf{(a)} SoftMol. \textbf{(b)} SoftMol (Unconstrained). DS across five targets under varying search budgets at $K_{\text{sample}}=8$. Results are averaged over 50 independent runs.}
    \label{fig:search_budget_radar}
\end{figure}

\textbf{Ablation of Gated MCTS.}
We conduct a comprehensive ablation study to isolate the impact of Gated MCTS configuration and SoftBD inference parameters, performing 360 experiments (9 variables $\times$ 4 settings $\times$ 5 targets $\times$ 2 models) by varying individual parameters from the default configuration (Table~\ref{tab:mcts_hyperparams}). Detailed analysis is provided in Appendix~\ref{app:ablation_details}.

\emph{Granularity and budget interaction.}
Figure~\ref{fig:ablation_granularity} investigates the interplay between soft-fragment granularity $K_{\text{sample}}$ and search budget $N_{\max}$.
Optimal granularity proves to be dynamic: under a strict budget ($N_{\max}=1000$), coarser fragments ($K_{\text{sample}}=8$) maximize efficiency by reaching valid chemical space with fewer steps; with increased compute ($N_{\max} \in \{5000, 10000\}$), finer granularity ($K_{\text{sample}}=4$) yields superior docking scores by enabling more precise local optimization.
Crucially, fixed extrema fail: fragment-level steps ($K_{\text{sample}}=12$) limit tree depth to 4--6 levels, restricting combinatorial complexity, whereas atom-level steps ($K_{\text{sample}}=2$) yield incomplete molecules even at $N_{\max}=10000$.
From the Hit Ratio perspective, SoftMol maintains near-perfect success rates ($>99\%$) across all settings, validating the effectiveness of the feasibility gate in confining the search to the drug-like manifold. In contrast, the unconstrained baseline exhibits significant degradation at extreme granularities ($K_{\text{sample}}=2$ and $K_{\text{sample}}=12$), confirming that the gate is essential for navigating the trade-off between affinity optimization and chemical validity.

\emph{Search budget scaling.}
Figure~\ref{fig:search_budget_radar} characterizes the compute--performance trade-off across search budgets spanning two orders of magnitude. Three trends emerge: \emph{(1) Monotonic improvement}: binding affinity increases with budget, with steepest gains at $N_{\max} \le 1000$; \emph{(2) Gradual saturation}: performance plateaus beyond $N_{\max}=5000$ at our default granularity ($K_{\text{sample}}=8$), indicating sufficient exploration of the current search space; notably, as suggested by Figure~\ref{fig:ablation_granularity}, employing a finer granularity (e.g., $K_{\text{sample}}=4$) could potentially unlock further gains by extending the optimization horizon given a larger budget; \emph{(3) Constraint cost}: the gap between configurations quantifies the affinity penalty of strict pharmacological compliance, which remains constant across budgets. Notably, even $N_{\max}=100$ exceeds many baselines in Table~\ref{tab:docking}, confirming the efficiency of the soft-fragment action space.

\section{Conclusion}
\label{sec:conclusion}

This work presents SoftMol, a unified framework for target-aware molecular generation that systematically co-designs representation, model architecture, and search strategy.
Central to this approach is the soft-fragment representation, a rule-free block formulation that enables diffusion-native modeling with tunable granularity.
Building on this foundation, the SoftBD architecture implements the first molecular block-diffusion language model, synergizing intra-block bidirectional denoising with inter-block autoregressive conditioning.
To ensure high-throughput sampling while simultaneously increasing structural validity, \emph{Adaptive Confidence Decoding} is integrated, while a gated MCTS mechanism explicitly decouples binding affinity optimization from drug-likeness constraints.
Empirically, SoftMol resolves the trade-off between generation quality and efficiency: it achieves 100\% chemical validity and a 6.6$\times$ speedup, while delivering a 9.7\% improvement in binding affinity and 2--3$\times$ higher diversity compared to state-of-the-art methods.

\clearpage
\section*{Impact Statement}
SoftMol reduces computational barriers in target-specific molecular design, potentially accelerating therapeutic development. However, this efficiency could be misused to design harmful compounds, as the model does not distinguish between therapeutic and toxic targets. Responsible deployment requires toxicity filters, ethical oversight, and controlled access to sensitive optimization objectives.


\bibliography{softmol}
\bibliographystyle{icml2026}

\clearpage
\appendix
\onecolumn

\section{Related Work}
\label{app:related_work}

\subsection{String-based Molecular Representations}
Molecular string representations have evolved from atom-level linearizations to more structured forms, encompassing syntax-constrained strings and rule-based fragmentation schemes. While SMILES~\citep{weininger1988smiles} serves as the standard linearization for its compactness, it lacks syntactic robustness, frequently leading to invalid valencies or unclosed rings in generative outputs~\citep{kusner2017gvae}. Although SELFIES~\citep{selfies2020} guarantees validity via a recursive derivation system, this robustness often compromises sequence compactness and interpretability.

To explicitly capture local chemical semantics, rule-based decomposition strategies have been widely adopted. Approaches such as SAFE~\citep{safe2024}, JT-VAE~\citep{jin2018jtvae}, and HierVAE~\citep{jin2020hiervae} decompose molecules into motif-level subgraphs (e.g., rings and linkers) based on chemical axioms like BRICS~\citep{degen2008brics} or RECAP~\citep{lew1998recap}. These methods ensure chemical validity of sub-components but suffer from rigid fragmentation rules and inflated sequence lengths due to auxiliary connectivity tokens (Figure~\ref{fig:paradigms}). Conversely, data-driven tokenization methods like SPE~\citep{spe} and t-SMILES~\citep{tsmiles2024} attempt to learn substructures statistically. However, their vocabularies remain static after training, lacking the flexibility to dynamically adjust granularity during inference.

Different from these discrete paradigms that depend on explicit chemical axioms or static vocabularies, we investigate a rule-free block representation strategy. By partitioning sequences into generic fixed-length segments, this approach removes the reliance on heuristic fragmentation and auxiliary connectivity tokens. This shifts the paradigm from rigid, rule-based preprocessing to a computation-centric representation, delegating the learning of chemical syntax and semantics to the generative model itself rather than enforcing it through static tokenization rules.

\subsection{Generative Models for Molecular Design}

\textbf{Autoregressive Models.} Autoregressive architectures represent the canonical framework for molecular string generation, wherein the joint distribution is factorized into a sequence of conditional probabilities. While high validity is achieved by Transformer-decoder architectures such as MolGPT~\citep{molgpt}, a fundamental inductive bias mismatch is introduced by their strictly causal factorization: molecules are intrinsically undirected graphs governed by bidirectional interactions rather than unidirectional sequence histories~\citep{jin2018jtvae, de2018molgan}. Consequently, the modeling of local chemical contexts is impeded, as the validity of topological substructures depends on their complete neighborhood~\citep{atz2021geometric, vignac2023}. Furthermore, a significant computational bottleneck is created by the $\mathcal{O}(L)$ latency inherent to sequential token-by-token generation.

\textbf{Diffusion Models.} Discrete diffusion constitutes a powerful non-autoregressive paradigm, enabling bidirectional context modeling and parallel generation. Although this framework has been adapted for molecular strings~\citep{genmol, mdlm}, a structural misalignment is introduced by the standard diffusion requirement for fixed-dimensional tensors. This constraint conflicts with the intrinsic nature of small molecules, which are discrete entities of variable length and strict topology~\citep{sterling2015zinc}. Consequently, computational efficiency is compromised by the necessity for extensive padding or complex length-modeling priors, which are suboptimal for the bounded and highly structured chemical space~\citep{kusner2017gvae, vignac2023}.

\textbf{Block Diffusion.} Block diffusion architectures~\citep{ssdlm,bd3lm} offer a hybrid solution by combining autoregressive block planning with parallel intra-block refinement. This paradigm is particularly well-suited for molecular generation, as it reconciles the need for variable-length generation (autoregressive) with the requirement for modeling rigorous local substructures (diffusion). By mitigating the limitations of pure AR and diffusion approaches, this hybrid framework enables efficient, structured generation without reliance on rigid fragmentation rules.

\subsection{Target-specific Molecular Design and Search}

\textbf{Reinforcement Learning and Genetic Algorithms.} Optimization in chemical space is frequently formulated as a search problem guided by generative models. Reinforcement learning (RL) strategies, notably REINVENT~\citep{reinvent}, cast molecule generation as a sequential decision process, fine-tuning policies via gradient-based updates. While potent, these methods require delicate scalarization of multi-objective rewards (combining affinities with pharmacological constraints) to avert mode collapse and syntactic degradation. This reliance often manifests as parameter sensitivity~\citep{goel2021molegular} or ``reward hacking,'' where models exploit scoring function artifacts to produce invalid high-scoring structures~\citep{tano2024reward,tom2025stereochemistry}. Alternatively, evolutionary algorithms like Graph-GA~\citep{graphga} apply heuristic mutation operators directly to molecular graphs. While sample-efficient, their exploration is inherently constrained to the local chemical neighborhood of the seed population.

\textbf{Fragment-Library and Retrieval-Based Methods.} A complementary paradigm restricts the search space to recombinations of a pre-existing fragment library. Frameworks such as GEAM~\citep{lee2024GEAM} and $f$-RAG~\citep{lee2024frag} construct candidates by assembling or retrieving motifs from large corpora. This approach significantly enhances sample efficiency and ensures local chemical validity. However, the achievable chemical variety is fundamentally bounded by the granularity and coverage of the underlying library: the step size is rigidly determined by the fragmentation scheme (e.g., BRICS), impeding the discovery of novel scaffolds absent from the predefined vocabulary.

\textbf{Monte Carlo Tree Search (MCTS).} MCTS offers a principled framework for balancing exploration and exploitation in sequential decision-making~\citep{chaslot2008monte}. In molecular design, ChemTS~\citep{chemts} pioneered token-level MCTS for SMILES, while MolDQN~\citep{moldqn} formulated optimization as a graph MDP. More recently, Trio~\citep{trio} introduced fragment-level expansion guided by a pretrained prior. However, existing methods face a granularity dilemma: atom-level search induces intractably deep trees, whereas rigid fragment-based MCTS restricts the reachable manifold. SoftMol addresses this by leveraging the soft-fragment representation to define a \emph{tunable, generative action space}. Our framework enables open-ended exploration with adjustable block granularity, while a \textbf{tunable feasibility gate} explicitly decouples constraint satisfaction from propery optimization, efficiently guiding search toward high-affinity regions without engaging computationally expensive docking for infeasible candidates.

\section{Attention Mask Design}
\label{app:attention_mask}

To enable structured learning across both corrupted and clean views of the input, we employ a custom block-aware attention scheme~\cite{bd3lm}. At each training step, we concatenate the noised sequence $\mathbf{x}_t$ and the clean sequence $\mathbf{x}$ into a single input of total length $2L$, then apply a hybrid attention pattern defined via a binary mask $\mathcal{M}_{\text{full}} \in \{0, 1\}^{2L \times 2L}$. 

The overall attention mask is decomposed into sub-components corresponding to specific dependency structures:
\begin{equation}
    \mathcal{M}_{\text{full}} =
    \begin{bmatrix}
        \mathcal{M}_{\text{BD}} & \mathcal{M}_{\text{OBC}} \\
        \mathbf{0} & \mathcal{M}_{\text{BC}}
    \end{bmatrix},
\end{equation}
where each component is defined as follows:
\begin{itemize}[leftmargin=*]
    \item \textbf{Block-Diagonal Mask}  ($\mathcal{M}_{\text{BD}}$): Enables bidirectional self-attention among tokens within the same block in the noised sequence $\mathbf{x}_t$. This facilitates local structure healing and refinement:
    \begin{equation*}
        [\mathcal{M}_{\text{BD}}]_{ij} = \begin{cases} 1 & \text{if } i, j \text{ belong to the same block in } \mathbf{x}_t, \\ 0 & \text{otherwise.} \end{cases}
    \end{equation*}

    \item \textbf{Offset Block-Causal Mask}  ($\mathcal{M}_{\text{OBC}}$): Allows each noised token in $\mathbf{x}_t$ to attend to tokens from strictly preceding blocks in the clean sequence $\mathbf{x}$, providing global autoregressive conditioning:
    \begin{equation*}
         [\mathcal{M}_{\text{OBC}}]_{ij} = \begin{cases} 1 & \text{if token } j \text{ is in a block strictly preceding that of } i, \\ 0 & \text{otherwise.} \end{cases}
    \end{equation*}

    \item \textbf{Block-Causal Mask}  ($\mathcal{M}_{\text{BC}}$): Maintains a standard causal mask over the clean sequence $\mathbf{x}$, ensuring that the autoregressive backbone representations are updated correctly:
    \begin{equation*}
        [\mathcal{M}_{\text{BC}}]_{ij} = \begin{cases} 1 & \text{if token } j \text{ is in the same or an earlier block as } i, \\ 0 & \text{otherwise.} \end{cases}
    \end{equation*}
\end{itemize}

\begin{figure}[t]
\begin{center}
\centerline{\includegraphics[width=1.0\columnwidth]{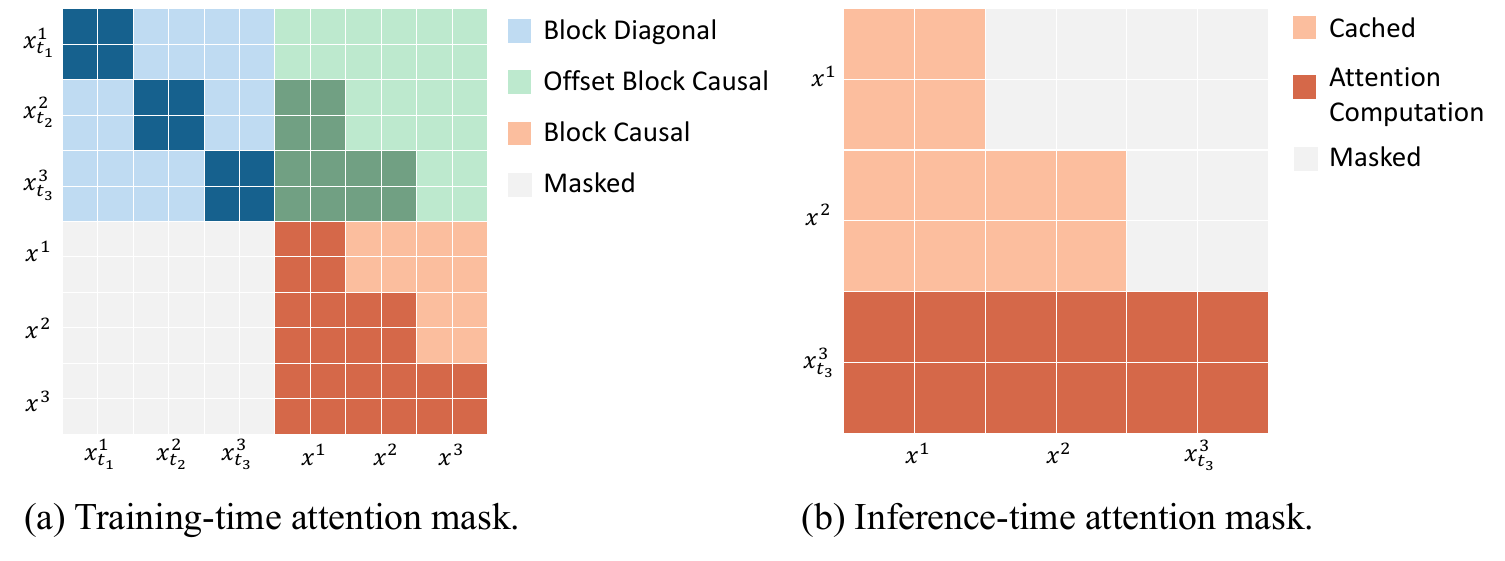}}
\caption{\textbf{Attention mask design.} \textbf{(a)} Training-time mask ($\mathcal{M}_{\text{full}}$) combining intra-block bidirectional attention ($\mathcal{M}_{\text{BD}}$), offset block-causal context ($\mathcal{M}_{\text{OBC}}$), and causal clean history ($\mathcal{M}_{\text{BC}}$). \textbf{(b)} Inference-time mask leveraging history caching for blocks $\mathbf{x}^{<b}$ while restricting active computation to the current bidirectional block $\mathbf{x}_t^b$.}
\label{fig:masked}
\end{center}
\vspace{-4mm}
\end{figure}

\paragraph{Inference-time Optimization.}
During inference, we adopt a simplified causal attention mechanism that reuses decoded blocks as a frozen prefix context. As illustrated in Figure~\ref{fig:masked} (b), previously generated blocks from history $\mathbf{x}^{<b}$ are cached to avoid redundant computation. Only the current noised block $\mathbf{x}_t^b$ is actively refined in each step; it attends bidirectionally within itself (similar to $\mathcal{M}_{\text{BD}}$) while attending causally to the unmasked tokens in the cached history. This design significantly reduces the memory footprint and computational cost by restricting active attention to the current block scale $\mathcal{O}(K_{\text{sample}}^2)$ rather than the full sequence length operations.

\section{Details of Training Objective}
\label{app:training_objective}

For completeness, we summarize the training objective used in SoftBD. While implementation operates in discrete time steps, we derive the objective in the continuous-time limit ($T \to \infty$) for theoretical completeness, following~\citet{mdlm,bd3lm}. We adopt the Negative Evidence Lower Bound (NELBO) formulation with the \emph{SUBS-parameterization}.

Consider a molecular sequence $\mathbf{x}$ partitioned into $B$ soft-fragments $\{\mathbf{x}^b\}_{b=1}^B$. The joint likelihood is factorized autoregressively over blocks:
\begin{equation}
    \log p_\theta(\mathbf{x}) = \sum_{b=1}^{B} \log p_\theta(\mathbf{x}^b \mid \mathbf{x}^{<b}).
    \label{eq:bd_factorization}
\end{equation}
Each conditional term $\log p_\theta(\mathbf{x}^b \mid \mathbf{x}^{<b})$ is modeled via a discrete diffusion process. By applying the standard variational bound and taking the continuous-time limit ($T \to \infty$), the objective can be decomposed into prior, reconstruction, and diffusion loss terms. 
Crucially, SoftMol utilizes the SUBS-parameterization \citep{bd3lm}, which enforces that unmasked tokens are never re-masked during the reverse process ($s < t$). Specifically, for any token index $i$ in block $b$, we set:
\begin{equation}
    p_\theta(\mathbf{x}_{s,i}^b = \mathbf{x}_{t,i}^b \mid \mathbf{x}_{t,i}^b \neq \texttt{[MASK]}) = 1.
\end{equation}
Under this constraint, the prior and reconstruction terms vanish.

The resulting objective simplifies to a weighted integral of the cross-entropy loss over time $t \in [0, 1]$:
\begin{equation}
    \mathcal{L}_{\text{BD}}(\mathbf{x}^b \mid \mathbf{x}^{<b}) = \mathbb{E}_{t \sim [0,1]} \mathbb{E}_{q(\mathbf{x}_t^b \mid \mathbf{x}^b)} \left[ -\frac{\alpha'_t}{1-\alpha_t} \mathcal{L}_{\text{CE}}(\mathbf{x}^b, p_\theta(\cdot \mid \mathbf{x}_t^b, \mathbf{x}^{<b})) \right],
\end{equation}
where $\alpha_t$ represents the noise schedule (probability of a token remaining unmasked, decaying from 1 to 0) and $\mathcal{L}_{\text{CE}}$ is the cross-entropy loss on the masked tokens ($x_{t,i}^b = \texttt{[MASK]}$).
The final training objective is the sum over all blocks:
\begin{equation}
    \mathcal{L}_{\text{BD}}(\mathbf{x}; \theta) = \sum_{b=1}^{B} \mathcal{L}_{\text{BD}}(\mathbf{x}^b \mid \mathbf{x}^{<b}).
\end{equation}
This formulation allows us to train the model to predict clean soft-fragments from their corrupted versions, conditioned on the clean history of previous blocks.

\section{Experimental Details}
\label{app:experimental_details}

\subsection{ZINC-Curated Dataset Preprocessing}
\label{app:dataset_details}
We constructed the ZINC-Curated dataset by collecting molecules from ZINC-22 \citep{sterling2015zinc} and applying a four-stage curation pipeline to ensure pharmaceutical relevance and structural diversity. The procedure, summarized in \Cref{alg:filtering}, proceeds as follows. First, we perform Physicochemical Filtering to discard molecules with QED $\le 0.5$ or SA $\ge 5$, corresponding to poor drug-likeness or low synthetic accessibility. Second, we enforce Structural Validity by removing compounds that contain undesirable elements such as Si or Sn, carry non-neutral charges, include free radicals, or exhibit overly complex topologies with more than two bridgehead atoms, rings larger than eight members, or more than ten rotatable bonds. We further exclude molecules with TPSA $> 140$ or known toxic or PAINS substructures. Third, we impose Medicinal Chemistry Rules via Lipinski's Rule of Five, constraining molecular weight to $100 \le \text{MW} \le 500$, lipophilicity to LogP $\le 5$, and hydrogen bond donor and acceptor counts. Finally, to mitigate dataset bias, we apply Diversity-Aware Stratification by grouping molecules by heavy-atom count between 4 and 49 and retaining only those whose Tanimoto similarity to previously accepted molecules in the same bucket is below 0.5.

\begin{algorithm}[t]
   \caption{ZINC-Curated curation pipeline}
   \label{alg:filtering}
\begin{algorithmic}[1]
   \STATE \textbf{Input:} Raw SMILES dataset $\mathcal{D}_{\text{raw}}$ (ZINC-22)
   \STATE \textbf{Output:} Curated dataset $\mathcal{D}_{\text{curated}}$
   \STATE Initialize buckets $\mathcal{H} \leftarrow \{ \mathcal{B}_4, \dots, \mathcal{B}_{49} \}$
   \STATE Define forbidden substructures $\mathcal{S}_{\text{ban}} \leftarrow \text{Toxic} \cup \text{PAINS} \cup \text{ForbiddenFragments}$
   
   \FOR{each molecule $m \in \mathcal{D}_{\text{raw}}$}
      \STATE \COMMENT{\textsc{1. Physicochemical filtering}}
      \IF{$\text{QED}(m) \le 0.5 \lor \text{SA}(m) \ge 5$}
         \STATE \textbf{continue}
      \ENDIF
      
      \STATE \COMMENT{\textsc{2. Structural validity}}
      \IF{$\text{Elements}(m) \cap \{\text{Si}, \text{Sn}\} \neq \emptyset \lor \text{Charge}(m) \neq 0 \lor \text{Radicals}(m) > 0$}
         \STATE \textbf{continue}
      \ENDIF
      \IF{$\text{Bridgehead}(m) > 2 \lor \text{MaxRing}(m) > 8 \lor \text{RotBonds}(m) > 10$}
         \STATE \textbf{continue}
      \ENDIF
      \IF{$\text{TPSA}(m) > 140 \lor \exists s \in \mathcal{S}_{\text{ban}} : s \subseteq m$}
         \STATE \textbf{continue}
      \ENDIF
      \IF{$\exists p \in \text{Atoms}(m, \text{P}) : \neg \text{HasSubstructure}(p, \texttt{[P](=O)(*)(*)}) $}
         \STATE \textbf{continue}
      \ENDIF

      \STATE \COMMENT{\textsc{3. Medicinal chemistry rules (Lipinski)}}
      \IF{$\text{LogP}(m) > 5 \lor \text{MW}(m) \notin [100, 500]$}
         \STATE \textbf{continue}
      \ENDIF
      \IF{$\text{HBD}(m) > 5 \lor \text{HBA}(m) > 10$}
         \STATE \textbf{continue}
      \ENDIF

      \STATE \COMMENT{\textsc{4. Diversity-aware stratification}}
      \STATE $h \leftarrow \text{HeavyAtomCount}(m)$
      \IF{$h \in [4, 49]$ \AND $\max_{m' \in \mathcal{B}_h} \text{Tanimoto}(m, m') < 0.5$}
         \STATE $\mathcal{B}_h \leftarrow \mathcal{B}_h \cup \{m\}$
      \ENDIF
   \ENDFOR
   \STATE $\mathcal{D}_{\text{curated}} \leftarrow \bigcup\limits_{h=4}^{49} \mathcal{B}_h$
\end{algorithmic}
\end{algorithm}

\subsection{Model Architecture}
\label{app:model_architecture}
SoftBD instantiates a discrete block-diffusion Transformer over soft-fragment tokens.
Concretely, we follow the DDiT backbone of BD3-LM\citep{bd3lm}: a stack of Transformer decoder blocks with masked self-attention over the concatenation of previously generated blocks and the current block, combined with position-wise feed-forward layers and residual connections.
The model vocabulary $\mathcal{V}$ consists of the atom-level SMILES character set from the training corpus and the control symbols $\{\texttt{[BOS]}, \texttt{[EOS]}, \texttt{[PAD]}, \texttt{[MASK]}\}$.
Input sequences are constructed by wrapping tokenized SMILES strings with start/end tokens as $(\texttt{[BOS]}, s_1, \dots, s_n, \texttt{[EOS]})$ and padding with \texttt{[PAD]} to a fixed length $L$ (e.g., $L=72$).
Soft-fragments are then obtained by slicing this full fixed-length tensor into blocks as described in Section~\ref{sec:softfragment}.
We tie the input token embeddings and output projection weights to reduce the parameter count and use dropout of 0.1 in all layers.

Across experiments, we use a family of SoftBD backbones ranging from 55M to 624M parameters.
All models use a 128-dimensional conditioning embedding to encode the diffusion timestep, and employ the scale-by-$\sigma$ parameterization from BD3-LM.
The 89M-parameter model is used for all main \emph{de novo} generation and target-specific molecular design experiments, while the 55M/72M models are used for corpus-sensitivity studies and the 116M/624M variants appear only in the scaling-law analysis in Section~\ref{sec:experiments}.
Table~\ref{tab:model_configs} summarizes the architectural hyperparameters of each backbone.

\begin{table}[h]
\caption{SoftBD backbone configurations used in this work. The maximum context length refers to the model's positional capacity; all experiments in this paper use SMILES sequences of length $L=72$.}
\label{tab:model_configs}
\begin{center}
\begin{small}
\begin{tabular}{ccccc}
\toprule
Params & Layers & Hidden size & Heads & Max context length \\
\midrule
55M    & 10     & 640         & 10    & 512 \\
72M    & 10     & 736         & 8     & 512 \\
89M    & 11     & 784         & 8     & 512 \\
116M   & 15     & 768         & 12    & 512 \\
624M   & 25     & 1408        & 16    & 1024 \\
\bottomrule
\end{tabular}
\end{small}
\end{center}
\vspace{-8mm}
\end{table}

\subsection{Training Details}
\label{app:training_details}
All SoftBD backbones in this work share the same optimization and training strategy.
We train with the AdamW optimizer (learning rate $3\times 10^{-4}$, $\beta_1=0.9$, $\beta_2=0.999$, $\epsilon=10^{-8}$, weight decay $=0$) and apply a linear warmup over the first 2{,}500 steps, after which the learning rate is kept constant.
We use mixed-precision training with bfloat16 activations and FP32 master weights via automatic mixed precision, and apply a dropout rate of 0.1 in all Transformer layers.
To stabilize training and reduce gradient noise, we maintain an exponential moving average (EMA) of the model parameters with decay $0.9999$ and employ antithetic sampling of diffusion noise.
We do not use gradient accumulation; all results are obtained with a single update per batch.

\subsection{Adaptive Confidence Decoding for SoftBD Details}
\label{app:efficient_inference}

We detail the algorithmic formulation of the three efficient inference mechanisms introduced in Section~\ref{sec:efficient_inference}. Algorithm~\ref{alg:inference} summarizes their integration into a unified generation pipeline.

\textbf{Analytic First-Hitting Schedule.}
Methodologically, First-Hitting Sampling acts as the temporal driver describing \emph{when} to sample. By sampling the next event time from the order statistics of the remaining mask count $m_t$, it determines strictly necessary evaluation points, skipping intervals where no tokens would be unmasked. In the algorithm, this corresponds to the time update step $t \leftarrow t \cdot u^{1/m}$, ensuring that every forward pass contributes to resolving at least one token.

\textbf{Greedy Confidence Decoding Implementation.}
At each denoising step, given probability tensor $\mathbf{P} \in \mathbb{R}^{N_{\text{batch}} \times K_{\text{sample}} \times |\mathcal{V}|}$ for the active block, we deterministically select the position-token pair with maximum confidence. Specifically, we compute the confidence $c_j = \max_{v} \mathbf{P}_{:, j, v}$ for each masked position $j$, then select $j^\star = \operatorname{argmax}_{j} c_j$ and reveal the corresponding argmax token. This ensures the model commits first to high-certainty substructures before resolving ambiguous features.

\textbf{Batched block-wise inference with caching.}
To enable efficient variable-length generation without memory overhead, we maintain a global sequence tensor $\mathbf{X}_{\text{accum}} \in \mathbb{R}^{N_{\text{batch}} \times L_{\text{max}}}$ updated in-place. For block $b$, we extract a sliding context window $[bK_{\text{sample}}-W, (b+1)K_{\text{sample}})$ from $\mathbf{X}_{\text{accum}}$, ensuring $O(W^2)$ attention complexity independent of total sequence length. To handle asynchronous termination, we track completion status per sequence: once an \texttt{[EOS]} token is generated, all subsequent positions in that sequence are frozen to \texttt{[EOS]} and excluded from updates. Generation terminates only when all sequences have produced at least one \texttt{[EOS]}, ensuring correct tensor alignment across variable-length molecules.

\begin{algorithm}[t]
\caption{Adaptive Confidence Decoding for SoftBD}
\label{alg:inference}
\begin{algorithmic}[1]
\STATE \textbf{Input:} Prior $p_\theta$, Batch size $N$, Block size $K_{\text{sample}}$, Context Window $W$, Max blocks $S$
\STATE \textbf{Output:} Generated molecules $\mathbf{X}_{\text{accum}}$
\STATE \textbf{Initialize:} $\mathbf{X}_{\text{accum}} \leftarrow \texttt{[BOS]} \times N$ (padded to max length)
\STATE \textbf{Initialize:} $\mathbf{m}_{\text{done}} \leftarrow \mathbf{0}_N$ (False)
\FOR{block $b = 0$ \TO $S-1$}
    \IF{$\forall i, \mathbf{m}_{\text{done}}^{(i)} = \text{True}$}
        \STATE \textbf{break}
    \ENDIF
    \STATE Define active block indices: $\mathcal{I}_{\text{act}} \leftarrow [\max(1, bK_{\text{sample}}), (b+1)K_{\text{sample}})$
    \STATE Define context window indices: $\mathcal{I}_{\text{ctx}} \leftarrow [\max(0, bK_{\text{sample}}-W), (b+1)K_{\text{sample}})$
    \STATE Initialize active block in $\mathbf{X}_{\text{accum}}$ with \texttt{[MASK]}
    \STATE Reset diffusion time per sequence: $\mathbf{t} \leftarrow \mathbf{1}_N$
    \LOOP
        \STATE Count masked tokens per sequence: $\mathbf{m} \leftarrow \sum_{k \in \mathcal{I}_{\text{act}}} \mathbb{I}(\mathbf{X}_{\text{accum}}[:, k] == \texttt{[MASK]})$
        \IF{$\mathbf{m} == 0$ (all resolved)}
            \STATE \textbf{break}
        \ENDIF
        \STATE \COMMENT{\textbf{1. Analytic First-Hitting Update}}
        \STATE Sample $\mathbf{u} \sim \mathcal{U}(0,1)^N$
        \STATE Update time: $\mathbf{t} \leftarrow \mathbf{t} \cdot \mathbf{u}^{1/\mathbf{m}}$
        \STATE \COMMENT{\textbf{2. Sliding Window Prediction}}
        \STATE Slice input: $\mathbf{X}_{\text{in}} \leftarrow \mathbf{X}_{\text{accum}}[:, \mathcal{I}_{\text{ctx}}]$
        \STATE Compute logits: $\mathbf{L} \leftarrow p_\theta(\mathbf{X}_{\text{in}}, \mathbf{t})$
        \STATE Extract probabilities for active block: $\mathbf{P} \leftarrow \text{Softmax}(\mathbf{L}_{:, -K_{\text{sample}}:, :})$
        \STATE \COMMENT{\textbf{3. Greedy Confidence Update}}
        \STATE Compute best confidence per pos: $\mathbf{C} \leftarrow \max_v \mathbf{P}$
        \STATE Mask already revealed positions in $\mathbf{C}$
        \STATE Identify target position: $j^\star \leftarrow \operatorname{argmax} \mathbf{C}$ (per sequence)
        \STATE Identify target token: $v^\star \leftarrow \operatorname{argmax} \mathbf{P}_{:, j^\star, :}$ (per sequence)
        \STATE Update tensor: $\mathbf{X}_{\text{accum}}[:, bK + j^\star] \leftarrow v^\star$
        \STATE \COMMENT{\textbf{4. Asynchronous Termination}}
        \IF{$\exists i: v^\star_i = \texttt{[EOS]}$}
            \STATE Update $\mathbf{m}_{\text{done}}$ for finished sequences
            \STATE Freeze remaining tokens in finished rows to \texttt{[EOS]}
        \ENDIF
    \ENDLOOP
\ENDFOR
\RETURN $\mathbf{X}_{\text{accum}}$
\end{algorithmic}
\end{algorithm}

\subsection{Hardware Infrastructure}
\label{app:hardware}
Computations are distributed across hardware configurations tailored to specific task requirements:
\begin{itemize}
    \item \textbf{Pretraining:} SoftBD models exceeding 116M parameters are trained on 8$\times$NVIDIA H100 GPUs, while smaller configurations ($\le$ 116M) utilize 8$\times$NVIDIA RTX 4090 GPUs.
    \item \textbf{\textit{De Novo} Generation:} All distribution learning evaluations and inference speed benchmarks (including all baselines) are strictly executed on a single NVIDIA H100 GPU to ensure a controlled comparison.
    \item \textbf{Target-Specific Optimization:} Tasks involving the full SoftMol framework (SoftBD generation coupled with Gated MCTS) leverage a cluster of NVIDIA RTX 4090 GPUs to support parallelized sampling and docking simulations.
\end{itemize}
Ablation studies adhere to this partition: generative configurations are evaluated on the single H100, while search parameter sweeps utilize the RTX 4090 cluster.

\subsection{Evaluation Protocols}
\label{app:evaluation_metrics}
We employ standardized metrics to evaluate molecular quality and diversity. Specifically, we use the RDKit library \citep{landrum2016rdkit} to obtain Morgan fingerprints, while measures of QED, SA, and diversity are calculated using the Therapeutics Data Commons (TDC) library \citep{huang2021tdc}. For binding affinity estimation, docking simulations are performed using QuickVina 2 \citep{alhossary2015qvina2}. All candidate molecules are converted to PDBQT format using OpenBabel prior to docking.

\subsection{\textit{De Novo} Generation Details}
\label{app:de_novo_details}

\textbf{Model Configuration.} We utilize the 89M-parameter SoftBD model trained on the ZINC-Curated dataset. The model was trained for 6 epochs, and we selected the checkpoint exhibiting the lowest validation loss for evaluation.

\textbf{Inference Strategy.} We employ our \textit{Adaptive Confidence Decoding} with a fine-grained inference setting of $K_{\text{sample}}=2$ to ensure high-fidelity generation. Sampling parameters are configured with nucleus sampling probability $p=0.95$ and temperature $\tau=0.9$ specifically for the efficiency analysis presented in Figure~\ref{fig:sampling_time}.

\textbf{Baselines.} When comparing against SAFE-GPT and GenMol, we use their official implementations and recommended hyperparameters to ensure fair reproduction.

\subsection{Target-specific Molecular Design Details}
\label{app:structure_based_details}

\textbf{Task Setup.}
We target five protein binding sites defined in the standard benchmark~\citep{rationaleRL}. The search space centers and box sizes are listed in Table~\ref{tab:docking_protocol}.
We adopt the median docking scores of known active molecules (Table~\ref{tab:docking_thresholds}) as strict baselines for defining high-affinity hits.

\textbf{MCTS Configuration.}
We employ the SoftMol search algorithm (Algorithm~\ref{alg:mcts}) utilizing the same SoftBD checkpoint as in Appendix~\ref{app:de_novo_details}. We use a coarse-grained expansion strategy ($K_{\text{sample}}=8$). As shown in Appendix~\ref{app:soft_fragment_length_details}, this setting offers an optimal trade-off: it is computationally cheaper than fine-grained generation while providing sufficient structural exploration for scaffold hopping.
Table~\ref{tab:mcts_hyperparams} lists the key hyperparameters.
We use a Children-Adaptive widening strategy where the expansion width dynamically scales based on the reward dispersion of valid children.

\textbf{Feasibility Gate \& Evaluation.}
During simulation, we apply a lightweight filter (QED $>0.5$, SA $<5.0$) before invoking the costly docking oracle. Candidates failing this gate receive a penalty score.
For the final evaluation, valid candidates are those that pass the gate, are unique SMILES, and successfully generate a 3D conformer via OpenBabel.

\textbf{Baselines.} To rigorously evaluate the performance of SoftMol, we differentiate between reported and reproduced baseline results:
\begin{itemize}[leftmargin=*, nosep]
    \item \textbf{Reported Results:} For the docking scores presented in Table~\ref{tab:docking}, we directly report the values from \citet{lee2024frag}. For the diversity results presented in Table~\ref{tab:circles}, we report the values from \citet{lee2024GEAM}. In both cases, we exclude GenMol as we reproduce it using the official implementation.
    \item \textbf{Reproduced Results:} For GenMol and all distributional analyses (Hit Ratio, DS Distribution in Figure~\ref{fig:combined_hit_rates}; Uniqueness in Figure~\ref{fig:uniqueness}), we fully reproduced the results using the official implementations (using $\delta=0.4$ for GenMol).
\end{itemize}
We generated 3,000 molecules per run across 3 independent runs for all reproduced baselines to ensure statistical significance.

\begin{table}[h]
    \caption{\textbf{Docking Experiment Protocol.} Grid box parameters for QuickVina 2 docking. All lengths are in Ångströms.}
    \label{tab:docking_protocol}
    \centering
    \small
    \begin{tabular}{cccc}
        \toprule
        \textbf{Target Protein} & \textbf{PDB Source / ID} & \textbf{Box Center} $(x, y, z)$ & \textbf{Box Size} $(x, y, z)$ \\
        \midrule
        parp1 & 4r6e & $(26.413, 11.282, 27.238)$ & $(18.521, 17.479, 19.995)$ \\
        fa7 & 1kl1 & $(10.131, 41.879, 32.097)$ & $(20.673, 20.198, 21.362)$ \\
        5ht1b & 4iar & $(-26.602, 5.277, 17.898)$ & $(22.500, 22.500, 22.500)$ \\
        braf & 3og7 & $(84.194, 6.949, -7.081)$ & $(22.032, 19.211, 14.106)$ \\
        jak2 & 3ugc & $(114.758, 65.496, 11.345)$ & $(19.033, 17.929, 20.283)$ \\
        \bottomrule
    \end{tabular}
\end{table}

\textbf{Hyperparameter Summary.}
Table~\ref{tab:mcts_hyperparams} summarizes the MCTS configuration used in target-specific molecular design experiments. The UCT selection mechanism employs exploration constant $C$ and weight $\lambda$ to balance average reward and peak performance. Search widths are controlled by a Children-Adaptive widening strategy: the root node initializes with $C_{\text{init}}$ children; non-root nodes use a base width $C_{\text{base}}$, which the adaptive mechanism dynamically adjusts between $C_{\min}$ and $C_{\max}$ based on the reward dispersion $I(s)$ and scaling factor $\beta$. Expansion generates $M$ candidate next-block soft-fragments per node using batched SoftBD sampling, while simulation completes $n_{\text{sim}}$ rollouts to assess node values.
\begin{table}[h]
    \caption{\textbf{Binding Affinity Thresholds.} The median docking scores (kcal/mol) of known active molecules used as thresholds for defining novel hits. Consistent with \citet{mood}, candidates with scores lower than these thresholds are considered high-affinity binders.}
    \label{tab:docking_thresholds}
    \centering
    \small
    \begin{tabular}{lc}
        \toprule
        \textbf{Target Protein} & \textbf{Threshold (kcal/mol)} \\
        \midrule
        parp1 & $-10.0$ \\
        fa7 & $-8.5$ \\
        5ht1b & $-8.8$ \\
        braf & $-10.3$ \\
        jak2 & $-9.1$ \\
        \bottomrule
    \end{tabular}
\end{table}

\begin{table}[h]
\caption{Hyperparameters for target-specific molecular design experiments.}
\label{tab:mcts_hyperparams}
\begin{center}
\begin{small}
\begin{tabular}{lc}
\toprule
Hyperparameter & Value \\
\midrule
\textbf{MCTS Parameters} & \\
Search budget ($N_{\text{max}}$) & 10000 \\
Exploration constant ($C$) & 2.1 \\
UCT weight ($\lambda$) & 0.5 \\
Children-Adaptive scaling ($\beta$) & 2 \\
Root initial width ($C_{\text{init}}$) & 20 \\
Base node width ($C_{\text{base}}$) & 8 \\
Minimum adaptive width ($C_{\min}$) & 8 \\
Maximum adaptive width ($C_{\max}$) & 10 \\
Batch expansion size ($M$) & 64 \\
Rollouts per expansion ($n_{\text{sim}}$) & 1 \\
Max tree depth ($D_{\text{max}}$) & 100 \\
\midrule
\textbf{Feasibility Gate} & \\
QED threshold ($\tau_{\text{QED}}$) & 0.5 \\
SA threshold ($\tau_{\text{SA}}$) & 5.0 \\
Unconstrained setting & $\tau_{\text{QED}}=0,\ \tau_{\text{SA}}=+\infty$ \\
\midrule
\textbf{SoftBD Sampling} & \\
Temperature ($\tau$) & 1.1 \\
Nucleus probability ($p$) & 1.0 \\
Soft-fragment length ($K_{\text{sample}}$) & 8 \\
Max context length ($L_{\text{max}}$) & 512 \\
Diffusion steps ($T$) & 128 \\
Random seeds & 42, 43, 44 \\
\bottomrule
\end{tabular}
\end{small}
\end{center}
\vspace{-4mm}
\end{table}

\subsection{Analysis of Soft-Fragment Length Details}
\label{app:soft_fragment_length_details}
For the granularity grid analysis in Section~\ref{app:effect_fragment}, we train independent 89M-parameter SoftBD models on ZINC-Curated ($L=72$) for 1 epoch and select the final checkpoint. This covers the full spectrum of $K_{\text{train}}, K_{\text{sample}} \in \{1, 2, 3, 4, 6, 8, 12, 24, 36, 72\}$, yielding 100 configurations. For each pair, we generate 1,000 molecules per run using nucleus sampling ($p=0.95$) and temperature $\tau=1.0$. All metrics (Validity, Uniqueness, Quality, Diversity, Docking-Filter, and Sampling Time) are averaged over 3 independent runs with seeds 42, 43, and 44. The resulting heatmaps (Figure~\ref{fig:robustness_heatmap}) visualize the full performance landscape.

\subsection{Ablation Studies Details}
\label{app:ablation_details}
\textbf{Adaptive Confidence Decoding (Table~\ref{tab:inference_ablation}).} We utilize the same fully-converged SoftBD checkpoint used in the \textit{de novo} (Appendix~\ref{app:de_novo_details}) and target-specific molecular design experiments. Inference is performed with $K_{\text{sample}}=8$, $L=72$, nucleus sampling $p=0.95$, and temperature $\tau=0.9$.

\textbf{Pretraining Corpus and Model Scaling (Table~\ref{tab:corpus_scaling}).}
All models presented in this ablation are trained for 1 epoch, and we select the final checkpoint for evaluation. Sampling is performed with $K_{\text{sample}}=8$, $L=72$, $p=0.95$, and $\tau=0.9$.

\textbf{Gated MCTS (Figure~\ref{fig:ablation_granularity}).}
We utilize the same checkpoint as in the target-specific molecular design experiments. Due to computational constraints, we generate 50 molecules per target for high-budget configurations ($N_{\max} \in \{5000, 10000\}$), while all other settings generate 100 molecules per target.

\section{SoftMol Search Pseudo-code}
\label{app:mcts_details}

Algorithm~\ref{alg:mcts} presents the complete SoftMol search procedure.
Each iteration follows the standard MCTS lifecycle:
\begin{itemize}[leftmargin=*,nosep]
\item \textbf{Selection:} Traverse the tree using UCT (Eq.~\ref{eq:uct}) to balance exploitation and exploration.
\item \textbf{Expansion:} Generate $M$ candidate soft-fragments via SoftBD ($K_{\text{sample}}=8$), filter duplicates, and instantiate a new child.
\item \textbf{Simulation:} Complete $n_{\text{sim}}$ rollouts to \texttt{[EOS]}. Assign $R_{\text{pen}}=-1.0$ for failures; otherwise, reward is $-\textsc{VinaScore}(\mathbf{x}, P)$.
\item \textbf{Backpropagation:} Update statistics $N(s)$, $\bar{R}(s)$, and $R^{\max}(s)$ along the path.
\end{itemize}

The root node initializes with $C_{\text{init}}=20$ children, while non-root nodes use a base width $C_{\text{base}}=8$ (adaptively widened up to $C_{\max}=10$). For SoftMol (Unconstrained), we set $(\tau_{\text{QED}}, \tau_{\text{SA}})=(0, +\infty)$ to bypass the feasibility gate and probe the affinity ceiling.

\begin{algorithm}[t]
   \caption{SoftMol target-specific optimization}
   \label{alg:mcts}
\begin{algorithmic}[1]
   \STATE \textbf{Input:} Protein $P$, SoftBD prior $p_\theta$, search budget $N_{\text{max}}$, batch size $M$, rollouts $n_{\text{sim}}$, search widths $(C_{\text{init}}, C_{\text{base}}, C_{\min}, C_{\max})$, feasibility gate $(\tau_{\text{QED}}, \tau_{\text{SA}})$, exploration params $(\lambda, C, \beta)$
   \STATE \textbf{Output:} Best molecule $\mathbf{x}^*$
   \STATE \COMMENT{Initialize root with $C_{\text{init}}$ children capacity}
   \STATE $root \leftarrow$ \text{CreateNode}(\texttt{[BOS]}, capacity $\leftarrow C_{\text{init}}$)
   \FOR{$i = 1$ \textbf{to} $N_{\text{max}}$}
      \STATE \COMMENT{Phase 1: Selection - traverse tree via UCT}
      \STATE $node \leftarrow root$, $expanded \leftarrow 0$
      \WHILE{\textbf{not} \text{IsTerminal}($node$) \textbf{and} $expanded = 0$}
         \STATE \text{UpdateSearchWidth}($node$, $C_{\min}$, $C_{\max}$) \COMMENT{Adaptive widening via Eq.~\ref{eq:adaptive_cap}}
         \IF{\textbf{not} \text{IsFullyExpanded}($node$)}
            \STATE \COMMENT{Phase 2: Expansion - sample and add child}
            \STATE $candidates \leftarrow$ \text{BatchSampleNextBlock}($p_\theta$, $node$, $M$) \COMMENT{Sample batch to ensure novelty}
            \STATE $candidates \leftarrow$ \text{FilterSiblingDuplicates}($candidates$, $node$)
            \STATE $child \leftarrow$ \text{CreateNode}(\text{RandomSelect}($candidates$), capacity $\leftarrow C_{\text{base}}$) \COMMENT{Select one to expand}
            \STATE \text{AddChild}($node$, $child$); $node \leftarrow child$
            \STATE $expanded \leftarrow 1$
         \ELSE
            \STATE $node \leftarrow$ \text{BestUCTChild}($node$) \COMMENT{Select via Eq.~\ref{eq:uct}}
            \IF{$node$ is \text{null}}
               \STATE \textbf{break}
            \ENDIF
         \ENDIF
      \ENDWHILE
      \STATE \COMMENT{Phase 3: Simulation - rollout and evaluate}
      \IF{$expanded = 1$}
         \STATE $reward \leftarrow -1.0$
         \FOR{$t = 1$ \textbf{to} $n_{\text{sim}}$}
            \STATE $\mathbf{x}_t \leftarrow$ \text{RolloutToEOS}($p_\theta$, $node$) \COMMENT{Complete to \texttt{[EOS]}}
            \STATE $reward_t \leftarrow -1.0$
            \IF{\text{IsValidSmiles}($\mathbf{x}_t$) \textbf{and} \text{PassFeasibilityGate}($\mathbf{x}_t, \tau_{\text{QED}}, \tau_{\text{SA}}$)}
               \STATE $reward_t \leftarrow -\text{VinaScore}(\mathbf{x}_t, P)$ \COMMENT{Negate for reward}
            \ENDIF
            \STATE $reward \leftarrow \max(reward, reward_t)$ \COMMENT{Take best rollout}
         \ENDFOR
      \ELSE
         \STATE $reward \leftarrow$ \text{CachedReward}($node$) \COMMENT{No expansion, use cached}
      \ENDIF
      \STATE \COMMENT{Phase 4: Backpropagation}
      \STATE \text{Backpropagate}($node$, $reward$) \COMMENT{Update $N, \bar{R}, R^{\max}$ to root}
   \ENDFOR
   \STATE \textbf{return} \text{BestRollout}($root$) \COMMENT{Return best molecule found}
\end{algorithmic}
\end{algorithm}

\section{Additional Experimental Results}
\label{app:additional_results}

\subsection{Robustness to Syntactic Discontinuities}
\label{app:prefix_completion}
Fixed-length segmentation inevitably partitions chemically meaningful substructures, potentially yielding unclosed branches, unmatched ring indices, or incomplete bracketed tokens. To evaluate the model's capacity for restoring such syntactic discontinuities, a controlled prefix-completion experiment was conducted. Generation was conditioned on a curated set of chemically broken SMILES prefixes designed to mimic characteristic boundary truncations.
These prefixes were selected to systematically evaluate the model's ability to recover various syntactic discontinuities, covering unclosed branches and rings, incomplete complex structures such as fused heterocycles, dangling bonds representing valency violations, and fine-grained stereo- or charge-related specifications. For each prefix, 1,000 completions were sampled across three independent runs. Validity was defined as parsability by RDKit, and strict adherence to the prefix constraint was verified via exact string matching.

As summarized in \Cref{tab:syntactic_robustness}, the model demonstrates near-perfect chemical validity across all distinct syntactic challenges. Despite the disruptive nature of fixed-length segmentation, the architecture reliably recovers local syntax and generates diverse, valid molecular continuations. It is observed that uniqueness naturally correlates with the specificity of the constraint; highly restrictive prefixes narrow the feasible chemical space, thereby reducing uniqueness while maintaining validity. Crucially, these results provide strong empirical evidence that the soft-fragment representation does not compromise molecular semantic integrity. Our SoftBD model effectively mitigates the arbitrary nature of fixed-length segmentation, ensuring that the intrinsic chemical logic remains intact.

\begin{table*}[t]
\caption{\textbf{Prefix completion from syntactically broken SMILES.} The model is conditioned on incomplete SMILES prefixes representing common truncation artifacts, including unclosed branches/rings, dangling bonds, and incomplete stereochemical/charge specifications. For each prefix, 1,000 samples are generated across three independent runs ($K_{\text{sample}} = 2$, $\tau = 1.0$, $p = 0.95$).}
\label{tab:syntactic_robustness}
\centering
\small
\resizebox{0.95\linewidth}{!}{%
\begin{tabular}{ll c c l}
\toprule
\textbf{Prefix} & \textbf{Syntactic Challenge} & \textbf{Validity (\%)} & \textbf{Uniqueness (\%)} & \textbf{Example Completion} \\
\midrule
\texttt{O=C(} & Unclosed Branch & \textbf{100.0 $\pm$ 0.0} & \textbf{99.6 $\pm$ 0.3} & \texttt{O=C(NCC=Cc1ccc(Br)cc1)c1cnn2c1OCCC2} \\
\texttt{C1CCCCC} & Unclosed Aliphatic Ring & \textbf{100.0 $\pm$ 0.0} & 99.1 $\pm$ 0.0 & \texttt{C1CCCCC1Nc1ccc(C\#N)cc1[N+](=O)[O-]} \\
\texttt{c1ccccc} & Unclosed Aromatic Ring & \textbf{100.0 $\pm$ 0.0} & 88.9 $\pm$ 1.3 & \texttt{c1ccccc1Nc1ccc(C2=NCCO2)cc1} \\
\texttt{c1cc(} & Branch within Aromatic System & \textbf{100.0 $\pm$ 0.0} & 94.2 $\pm$ 0.4 & \texttt{c1cc(NC[C@@H]2CCSC2)ncn1} \\

\texttt{C1CC1} & Small Ring Strain (3-mem) & 99.7 $\pm$ 0.1 & 85.4 $\pm$ 0.1 & \texttt{C1CC1N=C(O)c1cc(Br)ccc1O} \\
\texttt{c1nc2} & Unclosed Fused Heterocycle & 99.6 $\pm$ 0.1 & 74.3 $\pm$ 0.7 & \texttt{c1nc2c(c(=NC3CCSCC3)[nH]1)CCCC2} \\

\texttt{CC=} & Dangling Double Bond & \textbf{100.0 $\pm$ 0.0} & 96.2 $\pm$ 0.4 & \texttt{CC=CC(=O)c1ccc2c(c1)N=C(O)CCO2} \\
\texttt{CC\#} & Dangling Triple Bond & 99.5 $\pm$ 0.3 & 68.7 $\pm$ 1.2 & \texttt{CC\#N.COC(=O)c1ccc(F)cc1F} \\
\texttt{S(=O)(} & Hypervalent Sulfur State & \textbf{100.0 $\pm$ 0.0} & 74.6 $\pm$ 0.1 & \texttt{S(=O)(=O)NC1CCc2ccccc2C1} \\

\texttt{N[C@@H](} & Unclosed Stereocenter & \textbf{100.0 $\pm$ 0.0} & 63.9 $\pm$ 0.2 & \texttt{N[C@@H](CO)c1ccc2c(c1)N=C(O)CS2} \\
\texttt{c1ccc([N+](=O)} & Charge Balance \& Bracket Closure & \textbf{100.0 $\pm$ 0.0} & 48.7 $\pm$ 1.8 & \texttt{c1ccc([N+](=O)[O-])c(OC2CCCNC2)c1} \\
\bottomrule
\end{tabular}}
\end{table*}

\subsection{Analysis of Property Distributions}
\label{app:unfiltered_distributions}

\begin{figure*}[t]
    \centering
    \includegraphics[width=\textwidth]{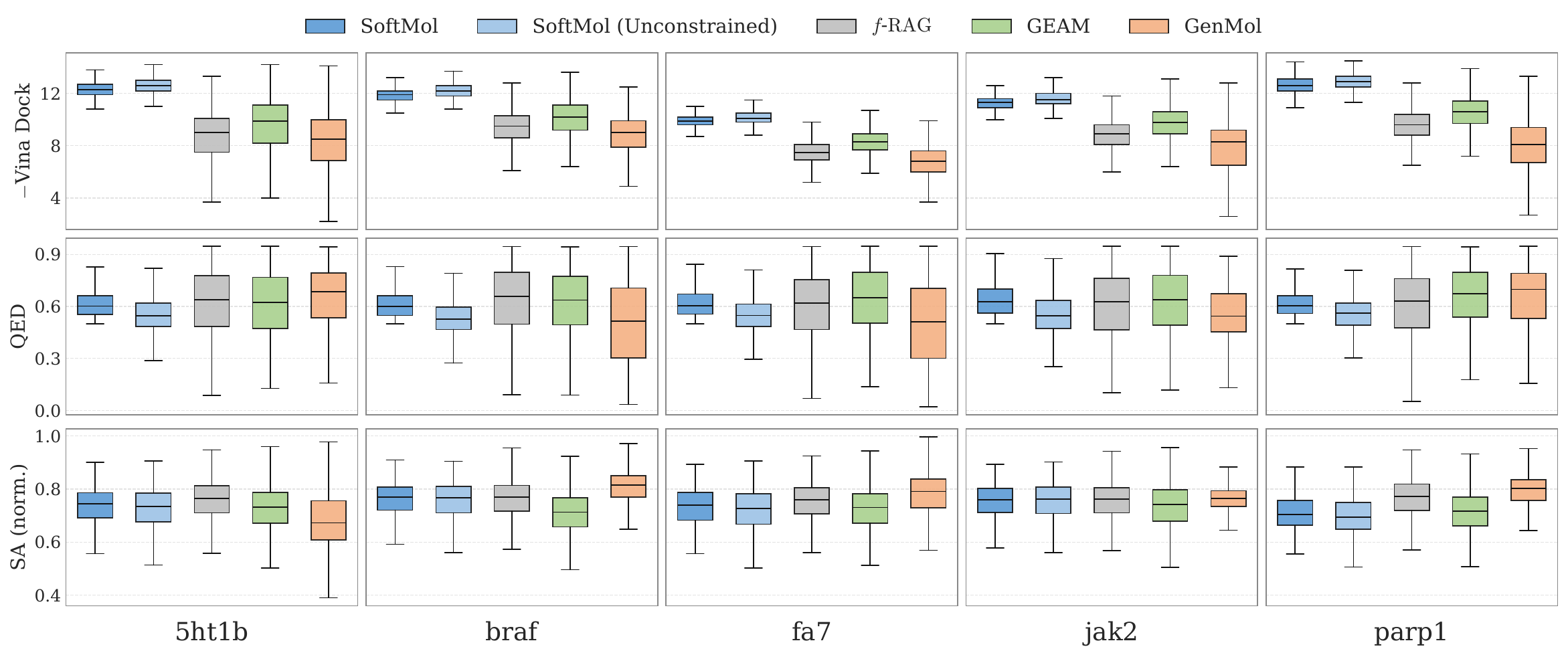}
    \caption{\textbf{Property Distributions.} Box plots of negative Vina Docking Score ($-$Vina Dock), QED, and normalized SA ((10-SA)/9) for unique molecules identified from 3,000 generated samples per method. SoftMol effectively constrains candidates to the high-quality drug-like region, whereas SoftMol (Unconstrained) pushes the affinity boundary.}
    \label{fig:unfiltered_dist}
\end{figure*}

To analyze the exploration characteristics of different methods, we compare the property distributions of the unique molecules derived from 3,000 generated samples per method in \Cref{fig:unfiltered_dist}.
As expected, SoftMol (Unconstrained) achieves the strongest binding affinities by exploiting the unconstrained search space.
However, SoftMol effectively concentrates the generated candidates within the drug-like manifold, resulting in distributions that align with the feasibility requirements.
In contrast, baselines such as GenMol exhibit significantly lower QED and SA scores, indicating a trade-off where affinity optimization compromises molecular quality.

\subsection{Ablation of diffusion budget $T$}
\label{app:decode_t}
We analyze the sensitivity of generation performance to the diffusion budget $T$ (\Cref{tab:decode_t_ablation}). Under the default efficient inference strategy, performance exhibits a sharp phase transition at $T = K_{\text{sample}}$: for $T < K_{\text{sample}}$, the budget is insufficient to resolve all block tokens, leading to validity collapse; for $T \ge K_{\text{sample}}$, metrics saturate immediately at near-optimal levels (Validity 100\%, Quality $\approx$ 81.9\%) with negligible variance. The First-Hitting mechanism effectively decouples computational cost from the budget $T$, maintaining low latency ($\approx$ 10\,s) even as $T$ increases.

In contrast, disabling our inference strategy reveals the limitations of standard diffusion decoding. While increasing $T$ gradually improves validity, it incurs a prohibitive linear computational cost. Crucially, even at extreme budgets ($T=5000$), the baseline fails to match the drug-likeness of the default strategy ($T=8$), plateauing at significantly lower Quality (57.0\% vs. 81.9\%) and Docking-Filter scores. This suggests that FH and GCD function not merely as accelerators, but as critical correction mechanisms that actively enforce chemical feasibility.

\begin{table*}[t]
\centering
\footnotesize
{\setlength{\tabcolsep}{4pt}%
\renewcommand{\arraystretch}{0.9}%
\begin{tabular}{lccccccc}
\toprule
Strategy & $T$ & Validity (\%) & Uniqueness (\%) & Quality (\%) & Docking-Filter (\%) & Diversity & Time (s) \\
\midrule
\multirow{7}{*}{\makecell[l]{Default\\(FH+GCD+Batch)}} 
      & 7     & 0.0 $\pm$ 0.0   & 0.0 $\pm$ 0.0 & 0.0 $\pm$ 0.0   & 0.0 $\pm$ 0.0 & 0.000 $\pm$ 0.000 & \textbf{8.922 $\pm$ 0.572} \\
      & 8     & \textbf{100.0 $\pm$ 0.0} & \textbf{100.0 $\pm$ 0.0} & \textbf{81.9 $\pm$ 2.0} & \textbf{98.5 $\pm$ 0.3} & 0.845 $\pm$ 0.001 & 9.490 $\pm$ 0.440 \\
      & 100   & \textbf{100.0 $\pm$ 0.0} & \textbf{100.0 $\pm$ 0.0} & \textbf{81.9 $\pm$ 2.0} & \textbf{98.5 $\pm$ 0.3} & 0.845 $\pm$ 0.001 & 9.864 $\pm$ 0.728 \\
      & 300   & \textbf{100.0 $\pm$ 0.0} & \textbf{100.0 $\pm$ 0.0} & \textbf{81.9 $\pm$ 2.0} & \textbf{98.5 $\pm$ 0.3} & 0.845 $\pm$ 0.001 & 10.766 $\pm$ 0.514 \\
      & 500   & \textbf{100.0 $\pm$ 0.0} & \textbf{100.0 $\pm$ 0.0} & \textbf{81.9 $\pm$ 2.0} & \textbf{98.5 $\pm$ 0.3} & 0.845 $\pm$ 0.001 & 10.928 $\pm$ 0.632 \\
      & 1000  & \textbf{100.0 $\pm$ 0.0} & \textbf{100.0 $\pm$ 0.0} & \textbf{81.9 $\pm$ 2.0} & \textbf{98.5 $\pm$ 0.3} & 0.845 $\pm$ 0.001 & 10.779 $\pm$ 0.877 \\
      & 3000  & \textbf{100.0 $\pm$ 0.0} & \textbf{100.0 $\pm$ 0.0} & \textbf{81.9 $\pm$ 2.0} & \textbf{98.5 $\pm$ 0.3} & 0.845 $\pm$ 0.001 & 11.157 $\pm$ 1.456 \\
      & 5000  & \textbf{100.0 $\pm$ 0.0} & \textbf{100.0 $\pm$ 0.0} & \textbf{81.9 $\pm$ 2.0} & \textbf{98.5 $\pm$ 0.3} & 0.845 $\pm$ 0.001 & 11.222 $\pm$ 0.660 \\
\midrule
\multirow{8}{*}{\makecell[l]{w/o FH+GCD\\(Batch)}} 
      & 7    & 9.7 $\pm$ 0.4 & \textbf{100.0 $\pm$ 0.0} & 6.2 $\pm$ 0.2 & 8.7 $\pm$ 0.1 & \textbf{0.876 $\pm$ 0.003} & 10.326 $\pm$ 0.497 \\
      & 8    & 14.9 $\pm$ 0.9 & \textbf{100.0 $\pm$ 0.0} & 9.5 $\pm$ 0.2 & 12.7 $\pm$ 0.6 & 0.874 $\pm$ 0.004 & 10.852 $\pm$ 0.895 \\
      & 100  & 86.2 $\pm$ 0.1 & \textbf{100.0 $\pm$ 0.0} & 52.6 $\pm$ 0.8 & 73.1 $\pm$ 0.4 & 0.855 $\pm$ 0.001 & 65.642 $\pm$ 0.991 \\
      & 300  & 89.1 $\pm$ 0.5 & \textbf{100.0 $\pm$ 0.0} & 55.8 $\pm$ 0.2 & 76.8 $\pm$ 0.4 & 0.855 $\pm$ 0.001 & 189.697 $\pm$ 0.411 \\
      & 500  & 89.8 $\pm$ 0.6 & \textbf{100.0 $\pm$ 0.0} & 55.0 $\pm$ 1.2 & 77.4 $\pm$ 1.4 & 0.855 $\pm$ 0.001 & 313.211 $\pm$ 0.863 \\
      & 1000 & 90.7 $\pm$ 0.4 & \textbf{100.0 $\pm$ 0.0} & 55.9 $\pm$ 1.1 & 77.9 $\pm$ 0.9 & 0.855 $\pm$ 0.000 & 623.491 $\pm$ 0.898 \\
      & 3000 & 90.3 $\pm$ 0.8 & \textbf{100.0 $\pm$ 0.0} & 56.9 $\pm$ 0.5 & 78.5 $\pm$ 0.3 & 0.855 $\pm$ 0.001 & 1726.290 $\pm$ 0.347 \\
      & 5000 & 90.5 $\pm$ 0.5 & \textbf{100.0 $\pm$ 0.0} & 57.0 $\pm$ 0.8 & 78.1 $\pm$ 0.8 & 0.854 $\pm$ 0.001 & 2476.522 $\pm$ 4.262 \\
\bottomrule
\end{tabular}
}
\vspace{2mm}
\caption{Effect of the diffusion budget $T$ under the default efficient inference strategy (\Cref{sec:efficient_inference}) and its ablation.
Results follow the ablation configuration (\Cref{app:ablation_details}) with 1,000 molecules.}
\label{tab:decode_t_ablation}
\vspace{-4mm}
\end{table*}


\subsection{Extended MCTS Ablations}
\label{app:detailed_ablations}

\begin{figure*}[p]
    \centering
    \includegraphics[width=\textwidth]{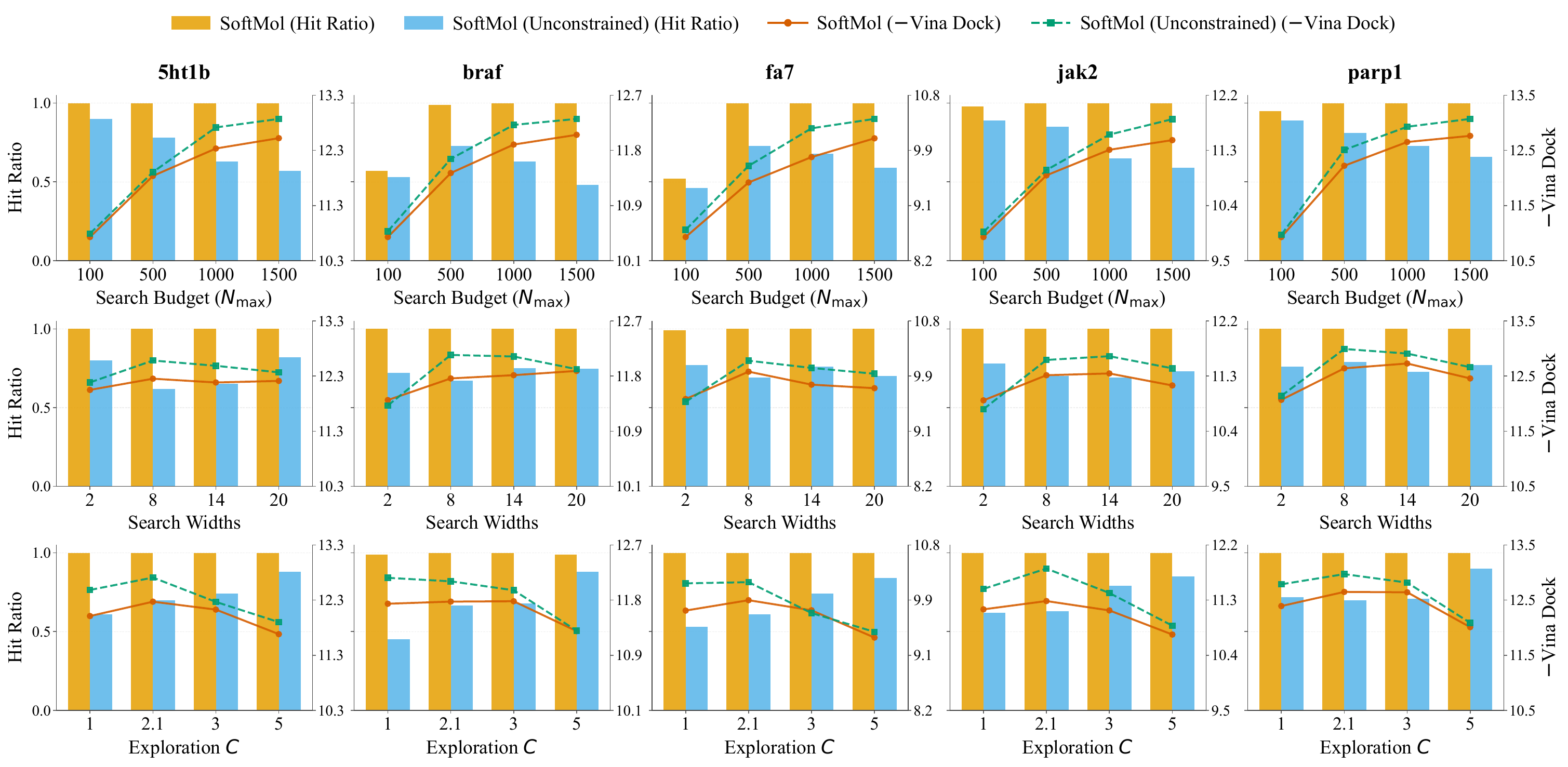}
    \caption{\textbf{Ablation on MCTS Hyperparameters.} Impact of Search Budget ($N_{\max}$), Node Expansion Width, and Exploration Constant $C$. Results are averaged across 5 targets from 100 independent runs.}
    \label{fig:ablation_mcts}
    \vspace{5mm}
    \includegraphics[width=\textwidth]{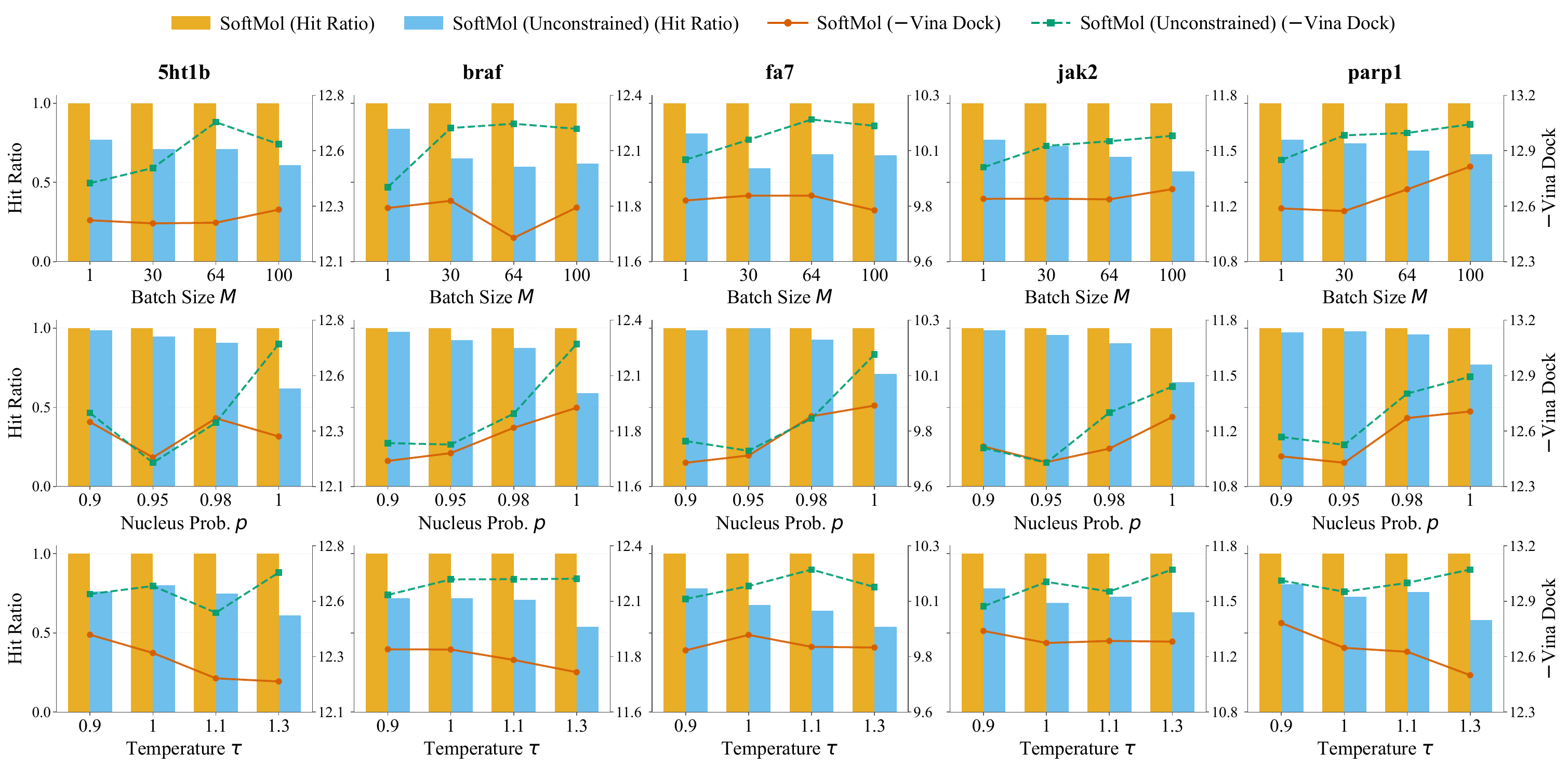}
    \caption{\textbf{Ablation on SoftBD Model Parameters.} Impact of Expansion Batch Size $M$, Nucleus Sampling Probability $p$, and Softmax Temperature $\tau$. Results are averaged across 5 targets from 100 independent runs.}
    \label{fig:ablation_model}
\end{figure*}

\textbf{MCTS Hyperparameters.}
Figure~\ref{fig:ablation_mcts} analyzes the search parameters.
Performance improves monotonically with search budget $N_{\max}$.
For node expansion width (candidates per node), a trade-off emerges: excessive width dilutes the search budget across shallow branches, while insufficient width limits exploration.
Similarly, the exploration constant $C$ balances exploring new chemotypes versus exploiting high-scoring branches; overly aggressive exploration ($C=5$) degrades performance by favoring uncertain, incomplete paths.

\textbf{SoftBD Model Configuration.}
Figure~\ref{fig:ablation_model} examines the generative prior.
Increasing the expansion batch size $M > 1$ enhances efficiency and diversity.
Disabling nucleus sampling ($p=1$) oddly improves performance, suggesting that truncating the distribution tail may hinder the discovery of strictly high-affinity novel structures.
Softmax temperature $\tau$ shows negligible impact, indicating the robustness of the pretrained prior.
Across all settings, SoftMol maintains a Hit Ratio near 100\%, validating the effectiveness of the feasibility gate.

\begin{figure*}[t]
    \centering
    \includegraphics[width=\textwidth]{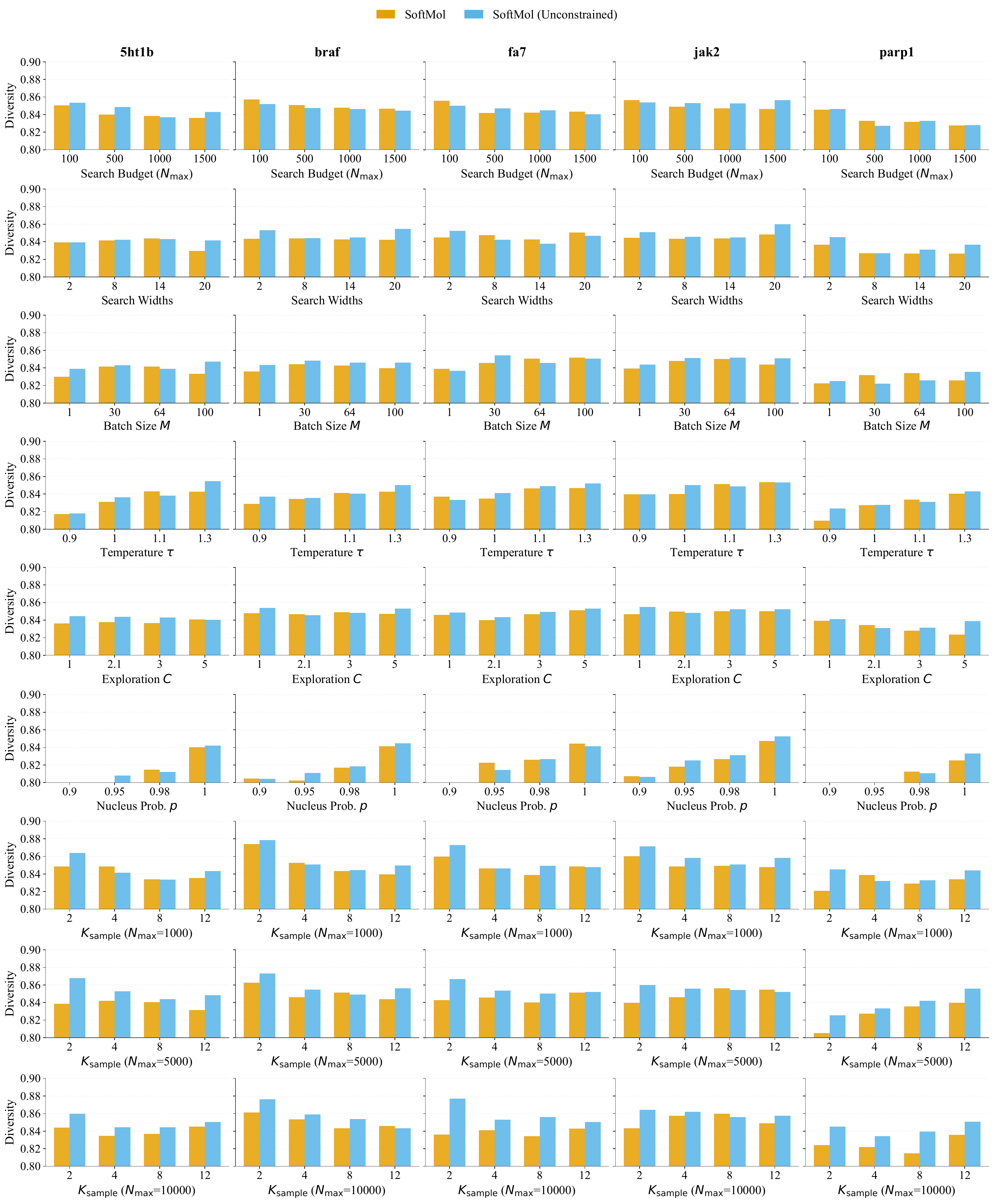}
    \caption{\textbf{Impact of Hyperparameters on Generative Diversity.} Internal diversity ($1 - \text{mean pairwise similarity}$) evaluated for SoftMol and the unconstrained baseline across five targets. Parameters are varied individually relative to the default configuration (Table~\ref{tab:mcts_hyperparams}). Results are averaged across 5 targets from 100 independent runs (50 runs for Search Budget $\in \{5000, 10000\}$ due to computational constraints).}
    \label{fig:ablation_diversity}
\end{figure*}

\textbf{Diversity Analysis.}
Figure~\ref{fig:ablation_diversity} quantifies the internal diversity of generated molecules across the full hyperparameter sweep. We first observe that SoftMol maintains high diversity comparable to the unconstrained baseline across all configurations, suggesting that pharmacophoric constraints do not significantly hinder chemical space exploration. Three key observations emerge. First, nucleus sampling probability $p$ exerts the strongest influence: reducing $p$ from 1.0 to 0.9 substantially increases diversity by broadening the generative sampling distribution, allowing exploration of lower-probability chemotypes. Second, temperature $\tau$ and batch size $M$ exhibit secondary effects: higher $\tau$ introduces stochasticity that diversifies outcomes, while larger $M$ amplifies the duplicate-filtering mechanism in batched expansion, directly increasing the fingerprint dissimilarity of instantiated children. Third, MCTS parameters show minimal impact: search budget $N_{\max}$, exploration constant $C$, and node widths yield near-constant diversity, confirming that molecular dissimilarity is primarily governed by the generative prior rather than search configuration. These results validate that SoftMol's batched generation mechanism effectively translates sampling stochasticity into molecular diversity, enabling the discovery of chemically dissimilar high-affinity candidates.

\clearpage
\section{Limitations}
\label{app:limitations}

While SoftMol presents a robust framework for target-aware molecular generation, we acknowledge the following limitations:

\textbf{(1) The 2D-3D Gap:} Our generative policy operates on 1D representations to maximize search throughput. Although we incorporate 3D docking during evaluation, the generation process itself lacks intrinsic awareness of 3D steric constraints. This may occasionally yield candidates that are topologically valid but conformationally energetically unfavorable within the binding pocket.

\textbf{(2) Reliance on Computational Proxies:} As with most \textit{in silico} benchmarks, we rely on Vina docking and heuristic scores (QED, SA) as optimization objectives. These are efficient but imperfect proxies for complex thermodynamic binding affinities and wet-lab synthetic feasibility.

\textbf{(3) Computational Cost and Search Budget:} While SoftBD achieves high inference speed, the full Gated MCTS procedure requires iterative rollout and oracle interaction. This represents a deliberate trade-off: investing higher computational cost per molecule to navigate the rugged energy landscapes of difficult protein targets.

\textbf{(4) Scope Limitations:} Our soft-fragment assumption is optimized for small drug-like molecules. Its effectiveness on macro-molecules or polymers, where structural dependencies span much larger logical distances, remains to be validated.

\end{CJK*}
\end{document}